\documentclass[11pt,a4paper]{article}
\usepackage{tabularx}

\newcommand\clearrow{\global\let\rowmac\relax}
\clearrow
\usepackage{multirow}
\usepackage{times,latexsym}
\usepackage{url}

\usepackage{breakurl}
\usepackage[breaklinks]{hyperref}
\usepackage[skip=0.25\baselineskip]{subcaption} % <-- new instruction
\usepackage[table]{xcolor}% http://ctan.org/pkg/xcolor
\usepackage[raggedrightboxes]{ragged2e}
\usepackage{arydshln}
\usepackage{capt-of}
\usepackage{pdfpages}

\usepackage[T1]{fontenc}
\usepackage{tikz}

\usetikzlibrary{tikzmark}

\usetikzlibrary{decorations,calc}
\pgfdeclaredecoration{simple line}{initial}{%<------https://tex.stackexchange.com/a/216086/197451
  \state{initial}[width=\pgfdecoratedpathlength-1sp]{\pgfmoveto{\pgfpointorigin}}
  \state{final}{\pgflineto{\pgfpointorigin}}
}
\tikzset{
   shift left/.style={decorate,decoration={simple line,raise=#1}},
   shift right/.style={decorate,decoration={simple line,raise=-1*#1}},
}
\usepackage{tikzscale}
\usetikzlibrary{arrows.meta}
\usetikzlibrary{bayesnet}
\usepackage{forest}
\usetikzlibrary{matrix}
\usetikzlibrary{decorations}
\usetikzlibrary{shapes.geometric,arrows,positioning}
\pgfdeclaredecoration{reserve}{initial}
{
  \state{initial}[width=\pgfdecoratedpathlength]
  {
    \pgfpathlineto{\pgfpoint{0pt}{-4pt}}
    \pgfpathlineto{\pgfpoint{0.15*\pgfdecoratedpathlength}{7pt}}
    \pgfpathlineto{\pgfpoint{0.2*\pgfdecoratedpathlength}{-7pt}}
    \pgfpathlineto{\pgfpoint{0.35*\pgfdecoratedpathlength}{5pt}}
    \pgfpathlineto{\pgfpoint{0.4*\pgfdecoratedpathlength}{-5pt}}
    \pgfpathcurveto{\pgfpoint{0.5*\pgfdecoratedpathlength}{3pt}}{\pgfpoint{0.6*\pgfdecoratedpathlength}{-3pt}}{\pgfpoint{0.6*\pgfdecoratedpathlength}{-3pt}}
    \pgfpathlineto{\pgfpoint{0.65*\pgfdecoratedpathlength}{6pt}}
    \pgfpathlineto{\pgfpoint{0.82*\pgfdecoratedpathlength}{-7pt}}
    \pgfpathlineto{\pgfpoint{\pgfdecoratedpathlength}{0pt}}
  }
  \state{final}
  {
    \pgfpathlineto{\pgfpointdecoratedpathlast}
  }
}
\tikzstyle{plate caption} = [caption, node distance=5pt, inner sep=0pt,
above left=10pt and -3pt of #1.north east] %
\usetikzlibrary{decorations.pathmorphing, patterns,shapes}

\usepackage{enumitem}
\usepackage{rotating}
\usepackage{pdflscape,array,booktabs}
\usepackage{soul}
\usepackage{comment}
\newcounter{statementnum}

\usetikzlibrary{shapes.geometric}
\usepackage[english]{babel}
\usepackage{amsthm}
\usepackage{amsmath}

\usepackage{thmtools}

\declaretheoremstyle[headfont=\normalfont\bfseries,bodyfont=\sffamily]{normalhead}
\declaretheorem[style=normalhead]{prompt}

\newtheorem{definition}{Definition}

\newlength{\qt}

\newcounter{rowcntr}[table]
\renewcommand{\therowcntr}{C-\arabic{rowcntr}}

\newcounter{aclcntr}[table]
\renewcommand{\theaclcntr}{ACL-\arabic{aclcntr}}

\newcounter{senatecntr}[table]
\renewcommand{\thesenatecntr}{SENATE-\arabic{senatecntr}}

\newcounter{designcntr}[table]
\renewcommand{\thesenatecntr}{DESIGN-\arabic{designcntr}}

\AtBeginEnvironment{tabular}{\setcounter{rowcntr}{0}}
\usepackage{wasysym}% provides \ocircle and \Box

\usepackage[acceptedWithA]{tacl2021v1}

\usepackage{xspace,mfirstuc,tabulary}

\newif\iftaclinstructions
\taclinstructionsfalse % AUTHORS: do NOT set this to true
\iftaclinstructions
\renewcommand{\confidential}{}
\renewcommand{\anonsubtext}{(No author info supplied here, for consistency with
TACL-submission anonymization requirements)}
\newcommand{\instr}
\fi

\iftaclpubformat % this "if" is set by the choice of options

\else

\fi

\title{Objectifying the Subjective: Cognitive Biases in Topic Interpretations}
\author{
  Swapnil Hingmire$^1$\Thanks{Work done at the University of Victoria, Canada.}\qquad Ze Shi Li$^{2*}$ \qquad Shiyu (Vivienne) Zeng$^2$ \qquad Ahmed Musa Awon$^2$ \AND Luiz Franciscatto Guerra$^2$ \qquad Neil Ernst$^2$
  \\ 
  $^1$Mehta Family School of Data Science and Artificial Intelligence,\\Department of Data Science, Indian Institute of Technology (IIT) Palakkad, Kerala, India \\
   \texttt{swapnilh@iitpkd.ac.in} \\
  $^2$Department of Computer Science, University of Victoria, Victoria, Canada\\
  \texttt{\{lize, shiyuzeng, ahmedmusa, luizguerra, nernst\}@uvic.ca} \\
}

\usepackage{forloop}

\newcounter{qr}

\usepackage{wasysym}
\usetikzlibrary{arrows,decorations.markings,plotmarks}

\makeatletter
\newcommand\newtag[2]{#1\def\@currentlabel{#1}\label{#2}}
\usepackage{array,ragged2e}
\newcolumntype{P}[1]{>{\RaggedRight\arraybackslash}p{#1}}
\makeatother

\begin{document}
\maketitle
\hypersetup{breaklinks=true}
\newcolumntype{N}{>{\refstepcounter{rowcntr}\therowcntr}c}
\newcolumntype{A}{>{\refstepcounter{aclcntr}\theaclcntr}c}
\newcolumntype{S}{>{\refstepcounter{senatecntr}\thesenatecntr}c}
\newcolumntype{D}{>{\refstepcounter{designcntr}\thedesigncntr}c}

\setlength{\abovecaptionskip}{0.5ex}
\setlength{\belowcaptionskip}{0.5ex}
\setlength{\floatsep}{0.5ex}
\setlength{\textfloatsep}{0.5ex}
 
\begin{abstract}
	Interpretation of topics is crucial for their downstream applications.  State-of-the-art evaluation measures of topic quality such as coherence and word intrusion do not measure how much a topic facilitates the exploration of a corpus. To design evaluation measures grounded on a task, and a population of users, we do user studies to understand how users interpret topics. We propose constructs of topic quality and ask users to assess them in the context of a topic and provide rationale behind evaluations. We use reflexive thematic analysis to identify themes of topic interpretations from rationales. Users interpret topics based on availability and representativeness heuristics rather than probability. We propose a theory of topic interpretation based on the anchoring-and-adjustment heuristic: users anchor on salient words and make semantic adjustments to arrive at an interpretation. Topic interpretation can be viewed as making a judgment under uncertainty by an ecologically rational user, and hence cognitive biases aware user models and evaluation frameworks are needed.
\end{abstract}

\section{Introduction}
Qualitative Content Analysis (QCA) is a dominant use case for topic models~\cite{DBLP:conf/nips/HoyleGHPBR21}. Topics, based on their most probable words ($T_W$), are used to discover new concepts, measure the prevalence of those concepts, assess causal effects, and make predictions~\cite{annurev:/content/journals/10.1146/annurev-polisci-053119-015921}. Examples of such use cases in social sciences are the study of culture~\cite{DIMAGGIO2013570}, politics~\cite{https://doi.org/10.1111/ajps.12103} and Islamophobia on social media~\cite{doi:10.1177/0957926516634546}. 

To speed up building and comparing new topic models objectively, several authors have proposed statistical coherence metrics ($C_{stat}$) (e.g., \citet{DBLP:conf/wsdm/RoderBH15}) that measure the connectedness of a meaning in $T_W$. Studies such as \citet{DBLP:conf/nips/HoyleGHPBR21} and \citet{ 10.1145/3617023.3617040} however question the validity of such metrics. 
\citet{DBLP:conf/nips/ChangBGWB09} proposed a word intrusion (WI) task based on identifying an ``intruder''  word inserted into a topic. A topic is coherent if it is easy to identify the intruder. WI, $C_{stat}$ and their extensions such as~\citet{Ying_Montgomery_Stewart_2022} and \citet{10.1162/coli_a_00518} implicitly assume that there is a unique, neutral, and objective perspective on coherence, and that this perspective is a single-dimensional assessment of the topic's interpretability. We question the validity of this assumption in the context of QCA.

\begin{table*}[!t]\footnotesize
	\begin{tabular}{|p{0.45cm}|p{3.75cm}|p{10.3cm}|}
		\hline
		\textbf{ID} & \textbf{Label} & \textbf{Highest probability words} \\ \hline
		\multicolumn{3}{|p{14cm}|}{\textbf{Immigration preferences}~\cite[Table 2]{Egami_Fong_Grimmer_Roberts_Stewart}} \\ \hline
		I6 & Crime, small amount of jail time, then deportation & enter \textbf{countri} \textbf{illeg} \textbf{person} \textit{jail} \textbf{deport} \textbf{time} proper \textit{imprison} determin \\  \hline
		I7 & No prison, deportation                             & \textbf{deport} \textit{prison} will \textbf{person} \textbf{countri} man \textbf{illeg} serv \textbf{time} sentence \\  \hline \hline
		\multicolumn{3}{|p{14cm}|}{ \textbf{ACL Anthology}~\cite[Table 2]{DBLP:conf/emnlp/HallJM08}} \\ \hline
		A5 & Categorial Grammar  & proof formula graph logic calculus axioms axiom theorem proofs lambek \\  \hline
		A7 & Classical MT                              & japanese method case sentence analysis english dictionary figure japan word \\  \hline
		A35 & Statistical MT                              & english word alignment language source target sentence machine bilingual mt \\  \hline  \hline
		\multicolumn{3}{|p{14cm}|}{ \textbf{arXiv submission in the Computing and Language section (cs.CL)}~\cite{MimnoArxiv2023_07}} \\ \hline
		C2 & - & logic quantum formal theory mathematical automata calculus proof finite regular \\ \hline \hline
		\multicolumn{3}{|p{14cm}|}{ \textbf{Poems from the ``Revising Ekphrasis'' corpus}~\cite{rhody2012topic}} \\ \hline
		P32 & - & night light moon stars day dark sun sleep sky wind \\ \hline 
	\end{tabular}
	\caption{Example topics from various studies}
	\label{topic_validation}
\end{table*}

Positionality (views and lived experiences shaped by identity and background~\cite{santy-etal-2023-nlpositionality}) and reflexivity (self-monitoring the impact of one's biases, beliefs, and personal experiences on their research) of researchers play an important role in the credibility of the findings of QCA~\cite{doi:10.1177/1468794112468475}. Furthermore, \citet{Hsieh_Shanon_QCA} define QCA as a method for \textit{the subjective interpretation of the content of text data}.
Neither subjectivity nor positionality nor reflexivity are considered in the generative process of topic modeling. They should be given more importance when using topics for QCA.

Consider topics A7 and A35 in Table~\ref{topic_validation}, originally from~\citet[Table 2]{DBLP:conf/emnlp/HallJM08} inferred on research articles in ACL Anthology. The authors labeled topic A7 as \textit{Classical MT} and topic A35 as \textit{Statistical MT}, with no explanation of how they labeled the topics. But based on these topic labels, they claimed a paradigm shift in machine translation. Without considering the positionality and reflexivity of the interpreters, these labels and the end conclusions can be contested.
Similarly, for topic A5, even though its $T_W$ does not contain the words ``Categorial'' and ``Grammar'', is labeled by the authors as ``Categorial Grammar''.

Positionality and reflexivity are important from the fairness, accountability, and transparency perspective, especially in dealing with texts in social sciences. Positionality can have a profound influence on interpreting \textit{contested} concepts (e.g., {freedom, democracy, war, genocide, abortion, hate crime}~\cite{Collier_contested_concepts}) or \textit{floating} concepts (mental health, race, gender, religion, and politics~\cite{Stuart_Hall_Race}). \citet{Wang_2023} support this argument: sociodemographics and positionality play a vital role in deciding ``safety'' of content generated by Conversational AI systems in terms of toxicity, harm, legal and health concerns, etc.

In this paper, we follow the view of \citet{10.1145/1390334.1390453}, who argues, in the context of information retrieval, that \textit{``If we can interpret a measure (...) in terms of an explicit user model (...), this can only improve our understanding of what exactly the measure is measuring''}. The (implicit) user model in metrics such WI and $C_{stat}$ is \textit{if a rational agent finds coherence in a topic's most probable words, then the agent will find the topic useful in all the circumstances}. This user model and metrics reward topic models that optimize topic  coherence but they cannot measure how much a topic is interpretable, i.e., facilitates the core tasks of exploration of a corpus or making inferences. Importantly, real users are only \textit{ecologically} rational and not \textit{axiomatically}~\cite{Gigerenzer2021}. Their knowledge, computational capacity, and environment constrain their decision-making, and hence are susceptible to cognitive biases. 

WI and $C_{stat}$ make strong and oversimplified assumptions of interpretability, do not account for the behavior of users, their reflexivity, positions, and experiences, and hence are not \textit{ecologically valid}\footnote{A model or metric is \textit{ecologically valid} if it accounts for the complexity of the underlying phenomenon and its associated human behavior in the real-world.}. There is a need to explicitly create a user model while evaluating topics.

\citet{DUPRET201349} argue that an evaluation metric should have two components: (i) A formal user model explaining the behavior of the population of users and (ii) A metric of performance based on the user model. The metric should have a strong positive correlation with a user's satisfaction in achieving their goal. 

As the first step in building a user model, in this paper, we report on user studies to understand how users interpret a topic in the first place. Following are the key contributions:

\begin{enumerate}[leftmargin=*,topsep=0pt]\itemsep-5pt
	\item Concepts of \textit{coherence} and \textit{interpretability} of topics are considered as slippery~\cite{DBLP:conf/nips/HoyleGHPBR21}. We show differences in these concepts and define them along multiple dimensions.
	\item We propose a set of constructs of topic quality in the context of QCA. We ask users to evaluate topics with respect to the constructs. 
	\item We do a QCA of topic labels and rationale using Reflexive Thematic Analysis (RTA)~\cite{Braun_Clarke_2012} to identify and analyze themes that shape topic interpretations.
\end{enumerate}

\section{Background and Related Work}\label{sec_background}
\textit{Coherence} is a metric of the \textit{interpretability} of topics~\cite{rahimi2023contextualized}, however often the two concepts are used interchangeably. We argue that coherence is a quality of words and focuses on their lexical or semantic relations. We adapt the definition of discourse coherence by~\citet{jbp_givon} for topic coherence:

\begin{definition}\label{def_coherence}
	\textit{Coherence of a topic is the continuity or recurrence of some element(s) across its most probable words. The elements are: (a) referents, (b) temporality, (c) aspectuality, (d) modality/mood, (e) location, (f) action/script }
\end{definition}

Interpretability on the other hand focuses on the semantic inference of words given the \textit{environment} of the user. Based on \cite{15ae9893-24f8-3286-b772-e96b33422e95,1957-01985-001}, we consider the environment as the user's physical and social circumstances and experiences that shape the positionality of the user. We adapt the definition of discourse interpretability by~\citet{Enkvist1990May01} for topic interpretability:

\begin{definition}\label{def_interpretablity}
	\textit{A topic is \textbf{interpretable} to those who can build around it a scenario or narrative explaining the situation and context in which these words are likely to be related to each other. The interpretation process is encapsulated in \textbf{a label}.
	}
\end{definition}    

\citet{Egami_Fong_Grimmer_Roberts_Stewart} apply topic models to open-ended responses from a survey on immigration preferences. Table~\ref{topic_validation} shows two example topics I6 and I7 from~\citet[Table 2]{Egami_Fong_Grimmer_Roberts_Stewart}. Through rigorous validation and discussions, they label the topics. The recurring theme in both the topics is \textit{immigration and deportation}. However, the authors' labels denote different actions: ``deportation \textit{with jail} vs deportation \textit{without jail}'' even though the topics share five words. Interpreting these topics merely on the most probable words while ignoring the authors' reflexivity and positionality can affect the experiments' results. If an intruder is inserted in both topics, then success in its identification will not help in their validation and labeling. Hence, we argue that WI focuses on the coherence of a topic and not on its interpretability. 

Several authors (e.g.,\cite{DBLP:conf/naacl/NewmanLGB10,DBLP:conf/nips/HoyleGHPBR21}) ask humans or LLMs such as GPT~\cite{DBLP:conf/eacl/RahimiMHNCA24} to rate coherence of a topic on a Likert scale (e.g., 1 to 3, where 1=useless or less coherent and 3=highly coherent). Such a scale, however, can be unintuitive for evaluators, leading to inconsistency between and within subjects. Moreover, using aggregation measures of numerical data to ordinal ratings is a fallacy~\cite{DBLP:conf/inlg/BelzK10}.

Such a rating is a \textit{revealed preference} (an agent's observed action) and not a \textit{normative preference} (the agent's actual interests) of a user. It does not capture how the topic was interpreted and why the specific rating is provided. The revealed preferences are incomplete and misleading measures of the normative preferences~\cite{BESHEARS20081787,Morewedge2023}. Rating based models and metrics ignore the \textit{inversion problem}~\cite{doi:10.1177/17456916231212138}: inverting mental states involved in topic interpretation from observed rating.

An important realization of a topic's interpretability is its label~\cite{doogan-buntine-2021-topic}. Several authors such as \citet{DBLP:conf/acl/LauGNB11,popa-rebedea-2021-bart} propose approaches for automatic labeling of topics. These approaches aim to identify ``one'' label that best captures semantic relations between $T_W$. 

\citet{JMLR:v18:17-069, doogan-buntine-2021-topic} assume an interpretable topic has a high agreement on labels. However, this assumption is not ecologically valid. For example, in Table~\ref{topic_validation}, the interpretation of topic C2 is subject to the user's exposure to ``Formal Semantics'' and ``Quantum Physics''. Hence, users may arrive at different labels, but that does not make the topic less or more interpretable. While applying LDA for Software Engineering artifacts, \citet{DBLP:journals/ese/HindleBZN15} make a similar observation: software developers and managers interpret topics differently; users find personally relevant topics easy to interpret. 

Moreover, a single labeling approach cannot be applied across domains. For example, topic C2 in Table~\ref{topic_validation} is about ``Formal Semantics'', but we cannot use similar reasoning and say that topic P32, inferred on ekphrastic poems, is about ``Celestial Phenomena'', as words in poems are often used in a figurative sense~\cite{rhody2012topic}. 

\citet{DBLP:conf/cikm/MaraniLB22} do human assessments of automatically generated labels of topics based on self-report questions and identify two latent dimensions of labels: \textit{Suitable} and \textit{Objectionable}. They focus on the end label provided by an algorithm, while we focus on \textit{how} a human labels a topic. Our proposed constructs and dimensions are broader and applicable across domains. \citet{Li:Mao:Stephens:Goel:Walpole:Fung:Dima:Boyd-Graber-2024} do simulations and user-based studies to compare topic models in an interactive task-based setting for QCA.  They do not discuss reflexivity, subjectivity and disagreement in topic interpretation and their effect on the end analysis. 

\citet{stammbach2023revisiting} and \citet{rahimi2023contextualized} use LLMs, to assess topic coherence and do intrusion tasks. LLM's response can be very sensitive to prompt~\cite{bubeck2023sparks}. Considering QCA as a use-case of topic models, there is a need to model \textit{the researchers, their knowledge, and the task}~\cite{Krippendorff2019} while employing LLMs. Moreover, given the adoption of LLMs for labeling topics as in \cite{rijcken2023towards}, we need more insight into potential pitfalls and biases.

\begin{table*}[!t]\scriptsize
	\begin{tabular}{NP{9.3cm}P{4.8cm}}
		\toprule
		\multicolumn{1}{c}{\textbf{ID}} & \textbf{Statement}                                                 & \textbf{Construct (Dimension)}  \\ \midrule
		\label{s_k2} & There is no jargon in this topic; even a novice can understand it ($\uparrow$)            & Knowledge  (Familiarity with domain) \\ \hline
		\label{s_int_srl2}  & I need to see the relevant documents to know what the topic is about ($\downarrow$)                 & Knowledge (Need for context) \\ \hline
		\label{s_c1} & The words of this topic have a connected meaning ($\uparrow$)                                     & Coherence (Continuity of a theme)\\ \hline
		\label{s_c3}  & This topic can be divided into subtopics for better understanding ($\downarrow$)                    & Coherence (Multiple sub-themes)   \\ \hline
		\label{s_int_srl1} & I can infer the situation and context in which these words are mentioned ($\uparrow$)                  & Interpretability (Communicative intent) \\ \hline
		\label{s_int_exm}  & I can quickly find a Wikipedia article to help someone understand this topic ($\uparrow$)         & Interpretability (Explicating the context) \\ \hline
		\label{s_int_ease} & This topic is easy to label ($\uparrow$)                                                            & Interpretability (Efforts)
		\\ \hline
		\label{s_int_priming}  & If I see this topic after some time, I don't think my label would change ($\uparrow$)      & Reflexivity (Time)   \\ \hline
		\label{s_int_pred} & Others will have similar interpretations ($\uparrow$)                                             & Reflexivity (Possible (dis)agreement) \\ \hline
		\label{s_int_order} & I will arrive at the same interpretation, even if the order of words is shuffled ($\uparrow$)  & Reflexivity (Order effects) \\ \bottomrule
	\end{tabular}
	\caption{Statements for user studies.  $\uparrow$ ($\downarrow$) indicates higher (lower) agreement with the statement implies better support for construct.}
	\label{statements_table}
\end{table*}

\subsection{Interpretability of Topics}
\begin{figure}[!h]
	\centering
	\includegraphics{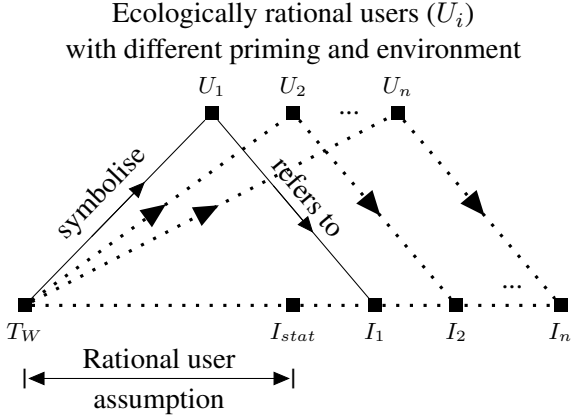}
	\caption{The Triangle of Reference. $T_W$: most probable words of a topic based on the generative and inferential assumptions of the topic model and the dataset.  $U_i$: Ecologically rational user ${i}$. $I_{stat}$: coherence and label of a topic based on statistical properties of $T_W$. $I_{i}$: $U_i$'s interpretation of topic.}
	\label{fig:triangle_reference}
\end{figure}

We created a conceptual schema to model the process of topic interpretation and labeling as the three corners of \textit{the triangle of reference} in Figure~\ref{fig:triangle_reference}, based on \citet{ogden1927meaning,DBLP:journals/aim/AroyoW15}.
In this schema, $T_W$ evokes a thought or reference as an idea in the mind of the user ($U_i$), based on their environment, priming, and goal of the analysis. $U_i$ arrives at an interpretation ($I_i$), typically as a label. The subjective interpretation is not directly connected with $T_W$, rather it is connected through the ecologically rational user. The triangle explains that for any given $T_W$, \textbf{many interpretations are possible}. 

State-of-the-art evaluation metrics are user agnostic; they focus only on the implicit link between $T_W$ and topic interpretation $I_{stat}$ (i.e., only on the base of the triangle) by making rational user assumptions.

\section{How Users Interpret and Label Topics}
We conduct an empirical study to examine user interpretation in terms of our triangle schema in Figure~\ref{fig:triangle_reference}, i.e., the impact of $U_i$ and $I_i$. We propose constructs for various aspects of topic interpretation: (i) {Knowledge}, (ii) {Coherence}, (iii) {Interpretability}, and (iv) {Reflexivity}.

\subsection{Constructs of Topic Quality}\label{sec_indicators}
We first show the 25 most probable words of a topic arranged vertically in the decreasing order of their probability. We do not show the probabilities associated with words, as users with no background of statistics behind topic models may find them unintuitive. We ask users their agreement with the statements in Table~\ref{statements_table} (``Strongly Disagree'' (score 0) to ``Strongly Agree'' (score 100))\footnote{Motivated by \citet{DBLP:conf/eacl/GattB10, DBLP:conf/emnlp/EthayarajhJ22} we use a continuous scale.}. We ask for a brief label for the topic. We also ask users to provide the detailed free-text rationale behind all their inputs to understand how the topics were interpreted. 

\begin{itemize}[leftmargin=*,topsep=0pt]\itemsep-5pt
	\item Statement~\ref{s_k2} captures the user's familiarity with the terminology of the underlying corpus.
	\item Statements~\ref{s_c1}, and~\ref{s_c3} capture multiple perspectives on coherence as per Definition~\ref{def_coherence}.
	\item Statements~\ref{s_int_srl1} and \ref{s_int_srl2} assess how much the user is able to think about possible communicative intent(s) that would have led to the formation of the topic.
	\item Statement~\ref{s_int_exm} is about finding an exemplar document: \textit{a narrative context in which the words have a clearer meaning}~\cite{BLAIR2002363}. 
	Finding such an article can help explicate the context that guides their inferences for QCA~\cite{Krippendorff2019}.     
	Statement~\ref{s_int_ease} is inspired from~\citet{doogan-buntine-2021-topic} about the difficulty in labeling a topic. 
	\item Statements~\ref{s_int_priming}, \ref{s_int_pred} and \ref{s_int_order} will help users to reflect on their interpretations in terms of time, disagreement, and word order respectively.
\end{itemize}

\subsection{Datasets}\label{sec_datasets}
Our first dataset is \textbf{ACL OCL} by~\citet{DBLP:conf/emnlp/RohatgiQAUK23} (\textit{ACL}) a scholarly corpus of papers hosted by the ACL Anthology published from 1952 to September 2022. As a second dataset, we consider \textbf{U.S. Senate Speeches} (\textit{SENATE}), the texts of U.S. Senate speeches provided by \citet{stanfordCongressionalRecord}. We focus on the speeches in the 114th session of Congress (2015-2017). 

The third dataset consists of Software Design (\textit{DESIGN}) related posts on StackOverflow (SO). Each SO post is allowed to have up to 5 tags that assigns the topic of the post. \citet{DBLP:conf/wcre/MahadiTE20} use 10 software-design related SO tags (viz. ``design-patterns'', ``software-design'', ``class-design'', ``design-principles'', ``system-design'', ``code-design'', ``api-design'', ``language-design'', ``dependency-injection'' and ``architecture'') to identify design-related posts on SO. We use SO API to extend the tags by querying tag descriptions (identified by PostTypes: TagWikiExcerpt and TagWiki) for the keyword ``design'' and having more than 100 posts. We manually removed tags that were not related to software design resulting in a set of 61 tags. We excluded UI design related tags such as css. We identified over 227 thousand design-related question-answer(s) pairs published till the end of December 2020 using the SOTorrent dataset~\cite{DBLP:conf/msr/BaltesT019}.

Appendix~\ref{app_pre_processing} provides details of the preprocessing of texts. We believe that based on the \textit{environments} of the users there will be QCA bias in interpretation of the texts and topics. 

\subsection{User Recruitment}
\citet{DBLP:conf/nips/HoyleGHPBR21} stress on evaluating topic models on a domain-specific corpus by people familiar with the domain. The central theme of this work is to do a detailed, case-oriented, and exploratory analysis of how users interpret topics, hence we focus on a limited number of users.

For the ACL dataset, we recruited users with exposure to NLP in terms of: (i) completing or teaching an NLP course, (ii) publishing an NLP research article, or (iii) working with natural language text. An overview of the task and an example was shared while recruiting users. We recruited 12 users (7 from Asia and 5 from North America) using convenience and snowball sampling strategies. 

For the SENATE dataset, it was difficult to recruit domain experts. We could recruit only four users from the USA through convenience and snowball sampling. We used Prolific\footnote{\url{https://www.prolific.com/} Last-accessed: 14-Jul-2025} to recruit 10 more users. Appendix~\ref{app_rec_senate_details} provides criteria for the recruitment. These users may not be experts in the field but are likely to have diverse opinions.

For the DESIGN dataset, we used Prolific to recruit 10 participants with exposure to Software Design. Appendix~\ref{app_rec_design_details} provides selection criteria for the recruitment.

\begin{table*}[!t] 
	\scriptsize 
	\begin{tabular}{lp{13.25cm}cc} 
		\toprule 
		\textbf{Topic ID} 	&	 \textbf{Highest probability words of topic}  	\tabularnewline \midrule		
		
		\multicolumn{2}{l}{\textit{\textbf{ACL dataset topics}} (\#Users: 12)} \tabularnewline  \midrule	
		\newtag{A1:Logic}{top_acl_1} 	&	  {{logical entailment semantics inference logic interpretation scope hypothesis predicate true predicates variable negation rule reasoning proof expression formula formal forms nli expressions variables operator premise}} 	 \tabularnewline \hline %2
		\newtag{A2:BERT}{top_acl_2}	&	  {{bert fine transformer tuning token shot pretrained loss roberta encoder devlin layer layers prediction pretraining downstream tuned masked embeddings transfer samples appendix batch span transformers}} 	  \tabularnewline \hline  %3
		\newtag{A3:Parsing}{top_acl_3}	&	  {{dependency parsing parser parse treebank tree head parsers trees dependencies structures constituent treebanks parses syntax gold arc pos penn transition attachment projective constituency constituents ccg}} 	  \tabularnewline \hline  %3
		\newtag{A4:Neural}{top_acl_4}	&	  {{neural layer network lstm embedding embeddings hidden encoder networks layers architecture deep cnn rnn mechanism prediction vectors decoder recurrent loss memory convolutional encoding softmax matrix}} 	  \tabularnewline \hline  %3
		\newtag{A5:Tweets}{top_acl_5}	&	  {{tweets social twitter user users media tweet posts comments messages post message online comment detection people hashtags thread email reddit forum emoji community privacy author}} 	 \tabularnewline  \midrule %3
		\multicolumn{2}{l}{\textit{\textbf{SENATE dataset topics}} (\#Users: 14)} \tabularnewline  \midrule	
		\newtag{S1:Students}{top_senate_1}	&	  {{students college education loan program student federal financial loans university colleges higher bill percent debt forprofit perkins universities school year proposed institutions banks street aid}} 	 \tabularnewline  \hline  %3
		\newtag{S2:Zika}{top_senate_2}	&	  {{health zika women planned parenthood virus funding care womens emergency public abortion bill control states services birth ebola disease centers babies pregnant state united cases}} 	  \tabularnewline \hline %3
		\newtag{S3:Freedom}{top_senate_3}	&	  {{freedom religious american history rights government united states americans human nation world america religion cuba liberty war state free faith nations democracy society political cuban}} 	  \tabularnewline \hline %3 
		\newtag{S4:Women}{top_senate_4}	&	  {{women rights act equal housing voting civil pay discrimination men law fair work americans maryland laws vote equality federal womens black state justice amendment country}} 	  \tabularnewline \hline %3
		\newtag{S5:Russia}{top_senate_5}	&	  {{united states world countries foreign international nations security russia russian country china ukraine allies europe national ambassador economic global policy region leadership government relations european}} 	  \tabularnewline \midrule
		\multicolumn{2}{l}{\textit{\textbf{DESIGN dataset topics}}  (\#Users: 10)} \tabularnewline  \midrule
		\newtag{D1:MVC}{top_design_1}	&	  {{model view controller mvc models views viewmodel data controllers partial mvvm logic create application rails presenter display pass app mvp pattern question django user viewmodels}} 	 \tabularnewline  \hline  %3
		\newtag{D2:Memory}{top_design_2}	&	  {{memory time performance cache size large number bit question stack load slow faster system times caching speed lot big usage case efficient fast data results}}  	 \tabularnewline \hline %3
		\newtag{D3:Database}{top_design_3}	&	  {{database table query sql tables data row record mysql column insert records rows queries update select columns stored key procedure create multiple server connection delete}} 	  \tabularnewline \hline %3 
		\newtag{D4:Threading}{top_design_4}	&	  {{lock thread locking transaction block read threads locks time locked write multiple process safe access wait shared prevent operation mutex release transactions update blocks case}} 	  \tabularnewline \hline %3
		\newtag{D5:Inheritance}{top_design_5}	&	  {{classes base parent child inheritance derived abstract subclass methods inherit extend method override subclasses virtual inherited extends superclass problem super add create inherits common extending}} 	  \tabularnewline \bottomrule
	\end{tabular} 
	\caption{Topics for user studies}
	\label{tab_user_study_topics}
\end{table*} 

\subsection{Inference and Interface}\label{sec_preprocessing}
\citet{DBLP:conf/emnlp/HoyleSGR22} show that LDA with Gibbs sampling (as implemented in MALLET~\cite{McCallumMALLET}) is more stable and reliable than newer neural models. Hence, we use MALLET to infer 50 topics on each dataset with default values of hyperparameters.  Table~\ref{tab_user_study_topics} lists the dataset-specific five topics for the user studies\footnote{We refer to each topic by an id based on the dataset, a number, and an informative word for ease of understanding.}. 

We customized the Potato text annotation tool by~\citet{DBLP:conf/emnlp/PeiAWZDSJ22} as an interface for the annotation task. Each user annotated five topics shown in random order. We expected 30 minutes for the study and paid each user the compensation of USD 19 for their participation\footnote{ The study was conducted with the approval of the IRB of our institute. Informed consent was obtained from all users.}. The average time for reading one topic, answering the agreement questions, and adding rationale, was 5.5 minutes.. 

All the user studies were carried out between 20 March 2024 and 15 April 2024. Our replication package is available for all the datasets, annotations, and Potato customization: \url{https://doi.org/10.5281/zenodo.14711182} Last-accessed: 14-Jul-2025.

\section{Quantitative Analysis}\label{sec_quant_analysis}
\begin{figure*}[!h]
	\centering
	\includegraphics[width=\textwidth]{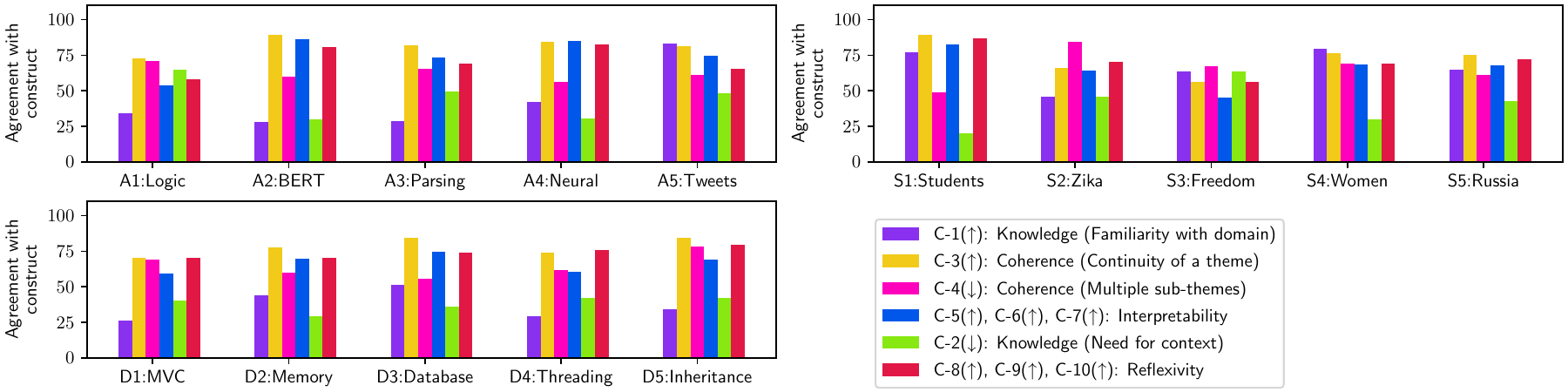}
	\caption{Plots for agreements of the users with the constructs. Appendix~\ref{app_sec_error_plots} provides topic-wise and construct-wise plots showing variations across users.}
	\label{fig_acl_senate_plots}
\end{figure*}

\begin{table*}[!t]\footnotesize
	\begin{tabular}{lllllllllllll}
		\toprule
		\textbf{Topic ID}  	&	\multicolumn{2}{c}{\textbf{User Reliability}} 	&	
		\multicolumn{10}{c}{\textbf{Coefficient of Variation (CoV)}} \\ \cmidrule(lr){2-3} \cmidrule(lr){4-13} 
		& \textbf{ICC} 	&	 \textbf{Interpretation} & \cellcolor{magenta!75}\ref{s_k2}	&	\ref{s_int_srl2}	&	\ref{s_c1}	&	\cellcolor{magenta!75}\ref{s_c3}	&	\ref{s_int_srl1}	&	\cellcolor{magenta!75}\ref{s_int_exm}	&	\ref{s_int_ease}	&	\ref{s_int_priming}	&	\ref{s_int_pred}	&	\cellcolor{magenta!75}\ref{s_int_order}\\ 
		& & & ($\uparrow$) & ($\downarrow$)  & ($\uparrow$)  & ($\downarrow$)  & ($\uparrow$)  & ($\uparrow$)  & ($\uparrow$)  & ($\uparrow$)  & ($\uparrow$)  & ($\uparrow$) \\ \midrule
		\ref{top_acl_1}    	&	0.56	&	 Moderate                	&	 \cellcolor{magenta!75}1.06 	&	 \cellcolor{magenta!35}0.43 	&	  \cellcolor{magenta!0}0.26 	&	 \cellcolor{magenta!35}0.5 	&	  \cellcolor{magenta!35}0.56 	&	 \cellcolor{magenta!35}0.67 	&	  \cellcolor{magenta!35}0.45 	&	 \cellcolor{magenta!35}0.54 	&	  \cellcolor{magenta!0}0.35 	&	 \cellcolor{magenta!35}0.45	\\
		\ref{top_acl_2}    	&	0.91	&	 Excellent               	&	 \cellcolor{magenta!75}1.18 	&	 \cellcolor{magenta!0}0.16 	&	  \cellcolor{magenta!0}0.13 	&	 \cellcolor{magenta!75}1.17 	&	  \cellcolor{magenta!0}0.3 	&	 \cellcolor{magenta!0}0.17 	&	  \cellcolor{magenta!0}0.19 	&	 \cellcolor{magenta!0}0.19 	&	  \cellcolor{magenta!0}0.36 	&	 \cellcolor{magenta!35}0.54	\\
		\ref{top_acl_3}    	&	0.72	&	 Moderate                	&	 \cellcolor{magenta!75}1.12 	&	 \cellcolor{magenta!0}0.31 	&	  \cellcolor{magenta!0}0.27 	&	 \cellcolor{magenta!35}0.78 	&	  \cellcolor{magenta!0}0.33 	&	 \cellcolor{magenta!35}0.5 	&	  \cellcolor{magenta!0}0.34 	&	 \cellcolor{magenta!35}0.56 	&	  \cellcolor{magenta!0}0.38 	&	 \cellcolor{magenta!35}0.45	\\
		\ref{top_acl_4}    	&	0.88	&	 Good                    	&	 \cellcolor{magenta!75}0.97 	&	 \cellcolor{magenta!0}0.22 	&	  \cellcolor{magenta!0}0.19 	&	 \cellcolor{magenta!75}1.16 	&	  \cellcolor{magenta!0}0.22 	&	 \cellcolor{magenta!0}0.22 	&	  \cellcolor{magenta!0}0.15 	&	 \cellcolor{magenta!0}0.21 	&	  \cellcolor{magenta!0}0.23 	&	 \cellcolor{magenta!35}0.67	\\
		\ref{top_acl_5}    	&	0.54	&	 Moderate                	&	 \cellcolor{magenta!0}0.32 	&	 \cellcolor{magenta!0}0.26 	&	  \cellcolor{magenta!0}0.25 	&	 \cellcolor{magenta!35}0.71 	&	  \cellcolor{magenta!0}0.29 	&	 \cellcolor{magenta!0}0.22 	&	  \cellcolor{magenta!35}0.41 	&	 \cellcolor{magenta!0}0.34 	&	  \cellcolor{magenta!35}0.66 	&	 \cellcolor{magenta!35}0.53	\\ \midrule
		\ref{top_senate_1} 	&	0.96	&	 Excellent               	&	 \cellcolor{magenta!0}0.24 	&	 \cellcolor{magenta!0}0.1 	&	  \cellcolor{magenta!0}0.11 	&	 \cellcolor{magenta!75}0.98 	&	  \cellcolor{magenta!0}0.28 	&	 \cellcolor{magenta!0}0.16 	&	  \cellcolor{magenta!0}0.13 	&	 \cellcolor{magenta!0}0.1 	&	  \cellcolor{magenta!0}0.11 	&	 \cellcolor{magenta!35}0.55	\\
		\ref{top_senate_2} 	&	0.76	&	 Good                    	&	 \cellcolor{magenta!35}0.61 	&	 \cellcolor{magenta!0}0.3 	&	  \cellcolor{magenta!0}0.38 	&	 \cellcolor{magenta!35}0.7 	&	  \cellcolor{magenta!0}0.34 	&	 \cellcolor{magenta!35}0.53 	&	  \cellcolor{magenta!0}0.35 	&	 \cellcolor{magenta!0}0.29 	&	  \cellcolor{magenta!0}0.35 	&	 \cellcolor{magenta!0}0.16	\\
		\ref{top_senate_3} 	&	0.45	&	 Poor                    	&	 \cellcolor{magenta!35}0.52 	&	 \cellcolor{magenta!35}0.5 	&	  \cellcolor{magenta!35}0.4 	&	 \cellcolor{magenta!0}0.35 	&	  \cellcolor{magenta!35}0.52 	&	 \cellcolor{magenta!35}0.61 	&	  \cellcolor{magenta!35}0.5 	&	 \cellcolor{magenta!35}0.4 	&	  \cellcolor{magenta!35}0.51 	&	 \cellcolor{magenta!0}0.28	\\
		\ref{top_senate_4} 	&	0.83	&	 Good                    	&	 \cellcolor{magenta!0}0.3 	&	 \cellcolor{magenta!0}0.22 	&	  \cellcolor{magenta!0}0.26 	&	 \cellcolor{magenta!75}0.85 	&	  \cellcolor{magenta!0}0.28 	&	 \cellcolor{magenta!35}0.41 	&	  \cellcolor{magenta!0}0.36 	&	 \cellcolor{magenta!0}0.25 	&	  \cellcolor{magenta!0}0.38 	&	 \cellcolor{magenta!0}0.37	\\
		\ref{top_senate_5} 	&	0.63	&	 Moderate                	&	 \cellcolor{magenta!35}0.42 	&	 \cellcolor{magenta!0}0.27 	&	  \cellcolor{magenta!0}0.22 	&	 \cellcolor{magenta!35}0.65 	&	  \cellcolor{magenta!0}0.32 	&	 \cellcolor{magenta!0}0.36 	&	  \cellcolor{magenta!0}0.25 	&	 \cellcolor{magenta!0}0.28 	&	  \cellcolor{magenta!0}0.29 	&	 \cellcolor{magenta!0}0.39	\\ \midrule
		\ref{top_design_1} 	&	0.72	&	 Moderate                	&	 \cellcolor{magenta!75}1.11 	&	 \cellcolor{magenta!35}0.46 	&	  \cellcolor{magenta!0}0.3 	&	 \cellcolor{magenta!75}0.95 	&	  \cellcolor{magenta!35}0.4 	&	 \cellcolor{magenta!35}0.49 	&	  \cellcolor{magenta!0}0.38 	&	 \cellcolor{magenta!0}0.21 	&	  \cellcolor{magenta!0}0.31 	&	 \cellcolor{magenta!0}0.31	\\
		\ref{top_design_2} 	&	0.78	&	 Good                    	&	 \cellcolor{magenta!35}0.65 	&	 \cellcolor{magenta!0}0.29 	&	  \cellcolor{magenta!0}0.25 	&	 \cellcolor{magenta!75}0.83 	&	  \cellcolor{magenta!0}0.2 	&	 \cellcolor{magenta!35}0.44 	&	  \cellcolor{magenta!0}0.35 	&	 \cellcolor{magenta!0}0.18 	&	  \cellcolor{magenta!35}0.43 	&	 \cellcolor{magenta!35}0.5	\\
		\ref{top_design_3} 	&	0.69	&	 Moderate                	&	 \cellcolor{magenta!35}0.69 	&	 \cellcolor{magenta!0}0.36 	&	  \cellcolor{magenta!0}0.18 	&	 \cellcolor{magenta!75}0.82 	&	  \cellcolor{magenta!0}0.28 	&	 \cellcolor{magenta!0}0.26 	&	  \cellcolor{magenta!0}0.39 	&	 \cellcolor{magenta!0}0.33 	&	  \cellcolor{magenta!35}0.43 	&	 \cellcolor{magenta!35}0.61	\\
		\ref{top_design_4} 	&	0.73	&	 Moderate                	&	 \cellcolor{magenta!75}1.1 	&	 \cellcolor{magenta!35}0.4 	&	  \cellcolor{magenta!0}0.26 	&	 \cellcolor{magenta!75}0.88 	&	  \cellcolor{magenta!35}0.43 	&	 \cellcolor{magenta!35}0.67 	&	  \cellcolor{magenta!0}0.28 	&	 \cellcolor{magenta!0}0.18 	&	  \cellcolor{magenta!0}0.39 	&	 \cellcolor{magenta!35}0.48	\\
		\ref{top_design_5} 	&	0.76	&	 Good                    	&	 \cellcolor{magenta!75}0.91 	&	 \cellcolor{magenta!0}0.37 	&	  \cellcolor{magenta!0}0.24 	&	 \cellcolor{magenta!75}0.86 	&	  \cellcolor{magenta!35}0.43 	&	 \cellcolor{magenta!35}0.58 	&	  \cellcolor{magenta!0}0.18 	&	 \cellcolor{magenta!0}0.33 	&	  \cellcolor{magenta!0}0.31 	&	 \cellcolor{magenta!0}0.2	\\ \bottomrule
	\end{tabular}
	\caption{Intraclass correlation coefficient (ICC) based index of inter-annotator agreement and its interpretation on the scale of poor to excellent~\cite{Koo2016-nz,2022-94730-001} and coefficient variation (CoV) for each topic-construct pair.}
	\label{tab_icc}
\end{table*}

Figure~\ref{fig_acl_senate_plots} shows the plots of the average agreement scores of users with multiple constructs. For ease of presentation, we plot the average of scores of the interpretability constructs (\ref{s_int_srl1}, \ref{s_int_exm}, and \ref{s_int_ease}) and those of the reflexivity constructs (\ref{s_int_priming}, \ref{s_int_pred}, and \ref{s_int_order}). 

\subsection{Topic Quality and Constructs}\label{sec_tq_constructs}
The following are the key observations from the agreement scores:
\begin{enumerate}[leftmargin=*,topsep=0pt,label=\textbf{O.\arabic*}]\itemsep-5pt %,wide = 0pt
	\item\label{tq_coherence} Users have contrasting views on coherence; they agree with \ref{s_c1}($\uparrow$) when they find a continuity of meaning but at the same time they find multiple sub-themes and ask to divide the topic (high agreement with \ref{s_c3}($\downarrow$)), especially in topics \ref{top_acl_1}, \ref{top_acl_3}, \ref{top_acl_5}; \ref{top_senate_2}, \ref{top_senate_4}, \ref{top_design_1}, and \ref{top_design_5}. 
    \item\label{tq_knowledge} Users argue that one needs to be more knowledgeable to understand topics--- \ref{top_acl_1}, \ref{top_acl_2}, \ref{top_acl_3}, \ref{top_senate_2}, \ref{top_design_1}, \ref{top_design_4}, and \ref{top_design_5}. 
    \item\label{tq_context} More context is required to interpret topics such as \ref{top_acl_1}, \ref{top_acl_3} and \ref{top_senate_3}. For \ref{top_acl_5}, users do not find any jargon, demand more context, and ask to divide the topic, but find continuity of a theme in $T_W$.
	\item\label{tq_reflex} Users show better reflexivity in their interpretation for topics \ref{top_acl_2}, \ref{top_acl_4}, \ref{top_senate_1}, \ref{top_design_4}, and \ref{top_design_5} compared to that of \ref{top_acl_1} and \ref{top_acl_3}.
	\item\label{tq_interpretability} The interpretability of topics \ref{top_acl_2}, \ref{top_acl_4}, \ref{top_senate_1}, and \ref{top_design_3} is relatively better than other topics in the respective datasets.
\end{enumerate}

These observations show that the constructs contribute differently in interpreting different topics. 

\begin{table*} \centering \footnotesize
	\begin{tabular}{l|llllllllll}
		\toprule
		&	\ref{s_k2} ($\uparrow$)	&	\ref{s_int_srl2} ($\downarrow$)	&	\ref{s_c1} ($\uparrow$)	&	\ref{s_c3} ($\downarrow$)	&	\ref{s_int_srl1} ($\uparrow$)	&	\ref{s_int_exm} ($\uparrow$)	&	\ref{s_int_ease} ($\uparrow$)	&	\ref{s_int_priming} ($\uparrow$)	&	\ref{s_int_pred} ($\uparrow$)	\\ \midrule
		\ref{s_int_srl2} ($\downarrow$)	&	-0.15	&		&		&		&		&		&		&		&				\\
		\ref{s_c1} ($\uparrow$)	&	-0.02	&	-0.66	&		&		&		&		&		&		&			\\
		\ref{s_c3} ($\downarrow$)	&	-0.27	&	0.51	&	-0.55	&		&		&		&		&		&				\\
		\ref{s_int_srl1} ($\uparrow$)	&	0.15	&	-0.73	&	\cellcolor{magenta!35}0.78	&	-0.37	&		&		&		&		&				\\
		\ref{s_int_exm} ($\uparrow$)	&	-0.19	&	-0.75	&	\cellcolor{magenta!35}0.88	&	-0.49	&	\cellcolor{magenta!35}0.78	&		&		&				&		\\
		\ref{s_int_ease} ($\uparrow$)	&	0.22	&	-0.68	&	\cellcolor{magenta!35}0.87	&	-0.62	&	\cellcolor{magenta!35}0.78	&	\cellcolor{magenta!35}0.84	&		&		&				\\
		\ref{s_int_priming} ($\uparrow$)	&	-0.13	&	-0.67	&	\cellcolor{magenta!35}0.82	&	-0.35	&	0.76	&	\cellcolor{magenta!35}0.85	&	\cellcolor{magenta!35}0.77	&		&				\\
		\ref{s_int_pred} ($\uparrow$)	&	0.07	&	\cellcolor{magenta!35}-0.9	&	0.64	&	-0.55	&	0.57	&	0.72	&	0.61	&	0.68	&				\\
		\ref{s_int_order} ($\uparrow$)	&	-0.21	&	-0.71	&	0.59	&	-0.25	&	0.62	&	0.65	&	0.56	&	\cellcolor{magenta!35}0.83	&	0.67			\\ 
		\bottomrule
	\end{tabular}
	\caption{Pairwise correlation between constructs based on the mean topic construct agreement across topics. Statistically significant correlations (with $p$-value $<$ .001) are highlighted.}
	\label{tab_corrleation_mean}
\end{table*}

\subsection{User Differences}\label{sec_icc}
We use Intraclass Correlation Coefficient (ICC)~\cite{Koo2016-nz,2022-94730-001} to study variations between the users' assessments of the constructs. ICC is a widely used coefficient to measure \textit{interrater reliability}. We use the \texttt{ICC} function from the \texttt{Psych} R library~\cite{PsyCh_Library} to employ ``Two-Way Random-Effects Model'' with ``Agreement'' to measure the reliability of the ``Average'' of scores of $k$-users (Table~\ref{tab_user_study_topics} provides the number of users for each dataset.). Table~\ref{tab_icc} provides the ICC statistic and its interpretation based on the guidelines by \citet{Koo2016-nz}.

We can observe that in the 8/15 topics, the reliability is \textit{Poor} or \textit{Moderate}, especially for the topics \ref{top_acl_1}, \ref{top_acl_5}, \ref{top_senate_3}, and \ref{top_senate_5}. To dive deeper in the user differences, we analyzed the coefficient of variation (CoV) among user scores for each topic. 

Table~\ref{tab_icc} provides the CoVs for each construct-topic pair, such that the higher the CoV, the more user differences. For the ease of understanding, we have highlighted the CoVs as: $(0, 0.4)$, $[0.4, 0.8)$, and $\geq 0.8$ as Low, \colorbox{magenta!35}{Medium}, and \colorbox{magenta!75}{High} respectively. We can observe that, the CoVs are highest for the constructs \textit{\ref{s_k2}:Knowledge (Familiarity with domain), \ref{s_int_exm}:Interpretability (Explicating the context)}, and \textit{\ref{s_int_order}:Reflexivity (Order effects)}. For almost all topics, the CoV for \textit{\ref{s_c3}: Coherence (Multiple sub-themes)} is very high, indicating that \textbf{users have different views on the coherence of the topics}. 

Table~\ref{tab_corrleation_mean} shows correlations between the constructs computed based on the mean construct agreement for each topic. There is a statistically significant correlation ($p$-value $<$ 0.001) among the constructs: \textit{\ref{s_c1}:Coherence (Continuity of a theme),} \textit{\ref{s_int_priming}:Reflexivity (Time)} and all the interpretability constructs (\textit{\ref{s_int_srl1}, \ref{s_int_exm}, \ref{s_int_ease}}). 

\textit{\ref{s_int_priming}:Reflexivity (Time)} is also strongly correlated with \textit{\ref{s_int_order}:Reflexivity (Order effects)}. There is a strong negative correlation between \textit{\ref{s_int_srl2}:Knowledge (Need for context)} and \textit{\ref{s_int_pred}:Reflexivity (Possible (dis)agreement)}. These constructs and correlations among them can be utilized in designing new evaluation framework.

Constructs \textit{\ref{s_int_srl2}:Knowledge (Need for context)} and \textit{\ref{s_k2}:Knowledge  (Familiarity with domain)} both capture \textit{Knowledge} requirements to interpret a topic, but there is no correlation among them. Similarly, coherence constructs \textit{\ref{s_c1}:Coherence (Continuity of a theme)} and \textit{\ref{s_c1}:Coherence (Multiple sub-themes)} have little correlation. Hence, items of the same construct may not be correlated to each other and are likely to capture its different aspects.

\subsection{LLM based Coherence}\label{sec_llm_coherence}
\begin{table*}[!t]\footnotesize \centering
	\begin{tabular}{l||ll|ll||ll|ll}
		\toprule
		& $\mathbf{C_{w/o}^3}$ & $\mathbf{R_{w/o}^3}$ & $\mathbf{C_{w/}^3}$ & $\mathbf{R_{w/}^3}$ & $\mathbf{C_{w/o}^5}$ & $\mathbf{R_{w/o}^5}$ & $\mathbf{C_{w/}^5}$ & $\mathbf{R_{w/}^5}$ \\ \toprule
		\rowcolor{gray!30} \ref{top_acl_1}	&	2.4	&	moderately related	&	2.6	&	very related	&	4	&	related	&	4	&	related	   \\	
		\ref{top_acl_2}	&	2	&	moderately related	&	2.2	&	moderately related	&	3.8	&	related	&	4.2	&	related	   \\	
		\rowcolor{magenta!10}\ref{top_acl_3}	&	2.8	&	very related	&	3	&	very related	&	4	&	related	&	4	&	related	   \\	
		\rowcolor{gray!30} \ref{top_acl_4}	&	2.4	&	moderately related	&	2.6	&	very related	&	4	&	related	&	4	&	related	   \\	
		\ref{top_acl_5}	&	2.8	&	very related	&	2.8	&	very related	&	4	&	related	&	4	&	related	   \\	  \midrule
		\ref{top_senate_1}	&	2.2	&	moderately related	&	3	&	very related	&	4	&	related	&	4	&	related	   \\	
		\ref{top_senate_2}	&	2	&	moderately related	&	2	&	moderately related	&	4	&	related	&	3.6	&	related	   \\	
		\ref{top_senate_3}	&	3	&	very related	&	3	&	very related	&	4.2	&	related	&	4.4	&	related	   \\	
		\ref{top_senate_4}	&	2.2	&	moderately related	&	2.8	&	very related	&	4	&	related	&	4	&	related	   \\	
		\rowcolor{magenta!10}\ref{top_senate_5}	&	3	&	very related	&	3	&	very related	&	4.2	&	related	&	4.8	&	very related	   \\	  \midrule
		\ref{top_design_1}	&	2.6	&	very related	&	3	&	very related	&	4	&	related	&	4	&	related	   \\	
		\ref{top_design_2}	&	2.2	&	moderately related	&	2.2	&	moderately related	&	4	&	related	&	4	&	related	   \\	
		\ref{top_design_3}	&	2	&	moderately related	&	2	&	moderately related	&	4	&	related	&	3.8	&	related	   \\	
		\ref{top_design_4}	&	2.4	&	moderately related	&	2.6	&	very related	&	4	&	related	&	4	&	related	   \\	
		\ref{top_design_5}	&	2.2	&	moderately related	&	2.8	&	very related	&	4	&	related	&	4	&	related	   \\	
		\bottomrule
	\end{tabular}
	\caption{Coherence scores: $C_{w/}^K$ ($C_{w/o}^K$) denote the LLM (GPT)-based mean coherence scores of a topic with (without) task and dataset description. Similarly, $R_{w/}^K$ ($R_{w/o}^K$) denote rating (mode Likert item). $K$ denotes the number of items in Likert scales; $K \in \{3,5\}$.}
	\label{tab_coherence}
\end{table*}

\begin{table*} \centering \footnotesize
	\begin{tabular}{lllllllllll}
		\toprule
		&	\ref{s_k2} ($\uparrow$)	&	\ref{s_int_srl2} ($\downarrow$)	&	\ref{s_c1} ($\uparrow$)	&	\ref{s_c3} ($\downarrow$)	&	\ref{s_int_srl1} ($\uparrow$)	&	\ref{s_int_exm} ($\uparrow$)	&	\ref{s_int_ease} ($\uparrow$)	&	\ref{s_int_priming} ($\uparrow$)	&	\ref{s_int_pred} ($\uparrow$)	&	\ref{s_int_order} ($\uparrow$) 	\\ \midrule
		$\mathbf{C_{w/o}^3}$	&	0.21	&	0.55	&	-0.43	&	-0.06	&	-0.52	&	-0.45	&	-0.24	&	-0.35	&	-0.51	&	-0.49	 \\ 
		$\mathbf{C_{w/}^3}$	&	0.26	&	0.2	&	-0.13	&	-0.1	&	-0.27	&	-0.3	&	-0.09	&	-0.01	&	-0.19	&	-0.14	 \\
		$\mathbf{C_{w/o}^5}$	&	0.41	&	0.44	&	-0.61	&	0.07	&	-0.62	&	-0.62	&	-0.42	&	-0.46	&	-0.44	&	-0.32	 \\
		$\mathbf{C_{w/}^5}$	&	0.21	&	0.16	&	-0.14	&	-0.26	&	-0.25	&	-0.2	&	-0.01	&	-0.04	&	-0.22	&	-0.11	 \\ 
		\bottomrule
	\end{tabular}
	\caption{Correlation between the constructs and coherence scores based on the mean construct agreement for each topic.}
	\label{tab_corrleation_coherence_mean}
\end{table*}

\citet{stammbach2023revisiting} claim that LLMs can assess the quality of topics. We verify their claim using their LLM-based coherence rating protocol. The protocol will also serve as a baseline that might represent a single, rational user. We apply prompts with ($C_{w/}^K$) and without ($C_{w/o}^K$) a task and dataset description and ask the LLM---GPT-3.5-Turbo to rate the word relatedness of a topic on the Likert scale ($K = 3$) from $1-3$ (``1'' = \textit{not very related}, ``2'' = \textit{moderately related}, ``3'' = \textit{very related}). 

To study the effect of the scale we repeated the experiments for a Likert scale ($K = 5$) of $1-5$ (``1'' = \textit{not very related}, ``2'' = \textit{somewhat related}, ``3'' = \textit{moderately related}, ``4'' = \textit{related}, ``5'' = \textit{very related}) with the same settings that of Likert scale ($K=3$). We prompt the LLM five times, with and without dataset details, for both scales, and report the mean score and rating (mode item of the scale). Table~\ref{tab_coherence} provides the coherence scores, and Appendix~\ref{app_llm_coherence} provides details of the prompts. 

The coherence scores show that subtleties in topic interpretations discussed in Sections~\ref{sec_tq_constructs} and \ref{sec_icc} are not captured by the LLM-based coherence ratings using both $R_{w/}^K$ and $R_{w/o}^K$ (for both the values of $K$), and hence their construct validity is limited. 

For example, the LLM assesses topic \ref{top_acl_3} 2.8 (\textit{very related}) and 3 (\textit{very related}), with $C_{w/o}^3$ and $C_{w/}^3$ respectively. However, according to \ref{tq_coherence}, \ref{tq_context}, and \ref{tq_reflex}, the topic has relatively poor average agreement scores by users. 

The LLM-based assessment for \ref{top_acl_1} and \ref{top_acl_4} is the same for $C_{w/o}^3$ and $C_{w/}^3$ (\textit{moderately related} and \textit{very related} respectively), however, as we can see in Figure~\ref{fig_acl_senate_plots} (and in Figure~\ref{fig_acl_senate_plots_app} from Appendix~\ref{app_sec_error_plots}) users find \ref{top_acl_4} more interpretable than \ref{top_acl_1} on all constructs. Moreover, ICC rating for \ref{top_acl_1} is \textit{Moderate}, while for \ref{top_acl_4} it is \textit{Good} (Table~\ref{tab_icc}).

The topics~\ref{top_senate_1}, \ref{top_senate_3}, \ref{top_senate_4} and \ref{top_senate_5} have exactly the same coherence rating ($R_{w/}^3$)---\textit{very related}, however, our observations in Section~\ref{sec_tq_constructs} (\ref{tq_coherence} to \ref{tq_interpretability}) show that users interpret these topics differently. Differences in ICC ratings for these topics, as shown in Table~\ref{tab_icc}, further strengthen this argument. A similar argument can be made in the context of the \textbf{ACL} and \textbf{DESIGN} topics. Moreover, the ratings of \ref{top_acl_3} and \ref{top_senate_5} are the same for $K=3$---\textit{very related}; however, we cannot conclude that the two topics from different domains have the same interpretability. 

Table~\ref{tab_corrleation_coherence_mean} shows the correlation between the coherence scores and the average agreement with a construct on 15 topics. We can observe that the correlations are not statistically significant. In Table~\ref{tab_coherence}, we can observe that there is little variation between the scores, despite the fact that the topics have different interpretations.

The scores and ratings does not help in assessing: (i) the utility of the topics for corpus exploration, (ii) (dis)similarity among the topics, (iii) variations in their interpretations, especially labels (Tables~\ref{tab_acl_labels}, \ref{tab_senate_labels}, and \ref{tab_design_labels} in Appendix~\ref{app_topic_labels} give topic labels provided by all the users.). 

\section{Reflexive Thematic Analysis (RTA)}\label{sec_themes}
\begin{table*}[!t]\footnotesize
	\begin{tabular}{P{1.25cm}P{2.5cm}P{11cm}}
		\toprule
		\textbf{ID} & \textbf{Topic label} & \textbf{Rationale} \\ \midrule 
		\multicolumn{3}{p{14cm}}{\textit{\textbf{A. Familiarity effects for AI-10}}} \\ \midrule
		Topic \ref{top_acl_2}  & 	Large language models &	the topic is easy to understand what it is "about", as it seems to capture relevant keywords to LLMs [...] \\ \midrule
		Topic \ref{top_acl_3} &	{linguistic} & 	\textbf{this topic for sure is less familar to me}. looking at top terms sicj as parser, tree, head, dependecy, I said ok it is about linguistic. but going further down the list the terms became less familiar to me.   [...] \\ \midrule\midrule
		\multicolumn{3}{p{14cm}}{\textit{\textbf{B. Context effect: Topic \ref{top_design_4}}}} \\ \midrule
		{DI-6}  & Operating Systems & Threads, mutex, locking, process are big hints. A subject I had in Uni was called "Operating Systems" that also covered this topic[...] \\ \midrule
		DI-2  &	Concurrency control, database managment and transaction processing.&	There is plenty of jargon, a novice without at least some background knowledge wouldn't understand the topic[...]I provided that label because the words such as "lock", "thread", "time", "multiple", and "transaction" reminded me of it the most[...]\\ \midrule\midrule
		\multicolumn{3}{p{14cm}}{\textit{\textbf{C. Memory based fallback effect: Topic \ref{top_acl_3}}}} \\  \midrule
		{AI-3} & {data} & key word: dependency, tree, treebank \textbf{Maybe data structure? I'm not sure}  \\ \midrule
		{AI-9} & {parser} & \textbf{Not familiar} with this topic, \textbf{but chose parser because I've used ``parser''}. \\ \midrule\midrule 
		\multicolumn{3}{p{14cm}}{\textit{\textbf{D. Ordering effect: Topic \ref{top_acl_1}}}} \\ \midrule 
		{AI-6} & {Entailment and Logical Reasoning} &  The topic is not easy to label because there seems to be a mixture of multiple topics like entailment (NLI) logical reasoning logic formulae and predicates[...] \textbf{The order of words is not playing much role in reducing the labelling confusion of this topic.} \\ \midrule
		{AI-8} &{Logical\newline entailment} & The first two words (logical entailment) suggested the topic for me and the other words seemed to align. \textbf{I might have gone for NLI if the term appeared at the top.} \\ 
		\bottomrule
	\end{tabular}
	\caption{Examples of topic interpretations showing the effect of differential salience of words. Table~\ref{tab_ecological_frequency_app} provides full texts of the interpretations.}
	\label{tab_ecological_frequency}
\end{table*}

We do a RTA of user-assigned topic labels and rationale to understand how topics are interpreted. The analysis involves six iterative phases of: \textit{(i) familiarisation; (ii) coding;  (iii) generating initial themes; (iv) reviewing and developing themes; (v) refining, defining, and naming themes; and (vi) writing up}~\cite{Braun_Clarke_2021}. The first author performed an RTA on 180 rationales and labels (5 interpretations of total 36 users)\footnote{Section~\ref{sec_positionality} provides the positionality of the authors.}.

This section discusses the major themes discovered using RTA. These themes capture different types of lexical and semantic inferences about possible communication intents in texts. 

\subsection{Salience}
Topic interpretations revolve around the salience of topic words as perceived by a user. We adapt the definition of word salience by \citet{Boswijk_salience} as \textit{the degree of prominence of a word, with respect to other words in $T_W$}. The salience of a word depends on factors such as: (i) Ecological Frequency (memory, recency, expertise), (ii) Familiarity in Context, and (iii) Word Position. 

\subsubsection{Ecological Frequency} A user determines the salience of a word based on the ease of retrieving experiences of the word in their memory. The ease of retrieval depends on ecological frequency~\cite{TVERSKY1973207, Anderson_Schooler}. Ecological frequency is subjective; it depends on the user's exposure to the word in their environment. Table~\ref{tab_ecological_frequency}-A, shows the interpretations of topics \ref{top_acl_2} and \ref{top_acl_3} by the user AI-10\footnote{We refer to a user by a unique id prefixed by dataset name, e.g. AI for ACL Interpreter.}. They \textit{know} the words in topic \ref{top_acl_2} and find the topic easier to interpret than topic \ref{top_acl_3}. %Some words and metaphors they represent have different meanings in different societies and cultures.

The context shapes the meaning of words; hence, selecting a particular context will affect a topic's interpretation. Table~\ref{tab_ecological_frequency}-B, shows DI-6 and DI-2 interpret Topic~\ref{top_design_4} in different contexts:  ``Operating Systems'' and ``Databases'' respectively. In literary texts, words and metaphors they represent have different meanings in different societies and cultures, and hence they may lead to different interpretations of topics~\cite{lakoff2008metaphors}.

\subsubsection{Familiarity in Context} The salience of a word also depends on how much a user sees a word in the context of $T_W$ and the domain of texts. This aspect differs from ecological frequency, as a meaning may be familiar in one context but not in the domain of interest. For example, consider interpretations of topic \ref{top_acl_3} in Table~\ref{tab_ecological_frequency}-C. Users AI-3 and AI-9 are not familiar with the topic. Their interpretations fall back on the concepts they are familiar with.  

\subsubsection{Word position}
Some users give importance to word order such that the top words influence interpretation as compared to the later words. Such an ``order effect'' is a commonly observed bias in interactions with search engines~\cite{conf/chiir/Azzopardi21}. Table~\ref{tab_ecological_frequency}-D gives examples of order effects seen in interpretations of topic \ref{top_acl_1} by AI-8. They would have changed their label if the word order had been different. AI-6 however does not give importance to the word order.

\begin{table*}[!t]\footnotesize
	\begin{tabular}{P{1.35cm}P{2.1cm}P{11.3cm}}
		\toprule
		\textbf{ID} & \textbf{Topic label} & \textbf{Rationale} \\ \midrule 
		{SI-4 for \textbf{\ref{top_senate_5}}} & {Ukraine and\newline  Russian\newline Conflict} & 	All the words have to do with countries that have some influence over \textbf{the current conflict between Russia and Ukraine}, as well as points of contention throughout the war[...] \textbf{This label does require some background knowledge on current events (so not everyone would come to the same conclusion)} [...] \\ \midrule
		SI-8 for \textbf{\ref{top_senate_5}} & Assistance & I labeled this section as assistance because ever since the Russians war against Ukraine has started the United states role has been providing assistance to Ukraine[...]\\ \midrule% \midrule 
		{SI-5 for \textbf{\ref{top_senate_1}}}&	{Student loan \newline discussions} & 	{Words like students, financial, loans, and banks make me think the speech was about \textbf{the recent} rates of student loans and how they need to be fixed.} \\ \midrule
		{SI-9 for \textbf{\ref{top_senate_2}}} & {Healthcare} & [...]women, abortion, planned, parenthood, control, babies, pregnant. All these words remind me on \textbf{the increasing} national regulation on women's healthcare rights.  \\ 
		\bottomrule
	\end{tabular}
	\caption{Example topic interpretations with presentism effect. Table~\ref{tab_biases_app} provides full texts of the interpretations.}
	\label{tab_biases}
\end{table*}

\subsection{Presentism or Projection Effects}\label{sec_presentism}
For the SENATE dataset topics, some users ignored the fact that the dataset contains speeches from 2015-2017 (which we informed them of). Instead, they are biased by more recent events\footnote{We refer the reader to Appendix~\ref{app_task_details} for more details}.
Such a bias, known as ``presentism'' (interpretation of the past in terms of the present), is observed in historical analysis~\cite{hunt2002against}. Table~\ref{tab_biases} gives an example of the presentism effect in the interpretation of topic \ref{top_senate_5} by user SI-4. %and SI-8. 
A possible bias is ``projection'' (overestimating the future based on the current state of mind).

According to the \textit{availability heuristic}~\cite{TVERSKY1973207}, users can find topics about contemporary and widely discussed events or issues (e.g., interpretations of topics \ref{top_senate_1} and \ref{top_senate_2} in Table~\ref{tab_biases} by users SI-5 and SI-9 respectively) or scientific paradigms~\cite{kuhn_ssr} easier to interpret. This may be the reason overall interpretability of topics \ref{top_acl_2}, \ref{top_acl_4}, \ref{top_senate_1}, \ref{top_senate_4}, and \ref{top_senate_5} is higher than other topics such as \ref{top_acl_1} and \ref{top_senate_2}. 

\subsection{Generalization and Stereotyping}
\begin{table*}[!t]\footnotesize
	\begin{tabular}{P{1.35cm}P{2.1cm}P{11.3cm}}
		\toprule
		\textbf{ID} & \textbf{Topic label} & \textbf{Rationale} \\ \midrule 
		\multicolumn{3}{p{14cm}}{\textit{\textbf{A. Stereotypes}}} \\ \midrule
		{SI-5 for\newline \textbf{\ref{top_senate_2}}} & {African Aid} & Things like, abortion, zika, ebola, are \textbf{all indicative of ongoing African societal and medical issues}. \textbf{This speech could be about aid to countries in Africa}. \\ \midrule
		{SI-9 for\newline \textbf{\ref{top_senate_3}}} & {International influence} & Unclear if these words are to represent the separation of church and state (freedom, religious, religion, faith) or \textbf{represent USA's involvement with spreading democracy into other countries} (Cuba, freedom, history, united, states, human, liberty, war, [...]\\ \midrule\midrule
		\multicolumn{3}{p{14cm}}{\textit{\textbf{B. Levels of Abstraction: Topic \ref{top_acl_2}}}} \\ \midrule 
		{AI-8} & {BERT} & This topic seems to be about transformers. \textbf{But the lack of GPT and other models}, and the rank of the word 'bert' led me to conclude this topic is actually about BERT. \\ \midrule% The associated terms such as roberts and masking further confirmed my analysis. \\ \midrule
		{AI-10} & {Large language models} & \textbf{the topic is easy to understand what it is "about", as it seems to capture relevant keywords to LLMs.} Seems the terms convey a cohesive topic [...] \\ \midrule\midrule 
		\multicolumn{3}{p{14cm}}{\textit{\textbf{C. Generalization for Topic \ref{top_acl_1}}}} \\ \midrule
		{AI-2} & {science} & {I think a person requires some experience in the sciences to understand these terms[...] I believe that this topic can be split into a topic that encompasses the mathematical[...] and another topic that focuses on logical reasoning[...]} \\  \midrule
		AI-9 & Logic & I can roughly guess the content because I've learned these concepts in a course, such as first-order logic, but I can't recall the name, as I used "logic" as the topic. \\ \midrule
		AI-4 & Natural language inference &	I am really hesitating between logics and nli as "entailment", "semantic", and "inference" are more related to NLI while the first word is "logical" \\ \bottomrule
	\end{tabular}
	\caption{Examples of topic interpretations showing the effect of Stereotyping and Generalization. Table~\ref{tab_stereotypes_app} provides full texts of the interpretations.}
	\label{tab_stereotypes}
\end{table*}

As limited context is available while interpreting a topic, users are likely to arrive at overgeneralized or stereotypical interpretations of certain words. Table~\ref{tab_stereotypes}-A gives examples of such interpretations. We refer the reader to \cite{Dionne20208} for discussion on stereotypes about ``Africa'', ``Zika'' and ``Ebola'', and to \cite{Bernell2012} for discussion on some of the perceptions of Cuba by the US residents. We argue that such stereotypical interpretations are likely to happen in the case of contested concepts~\cite{Weber_2009}. 

Topic labels vary across the spectrum of maximum to minimum levels of abstractions (LoA) of a given $T_W$, leading to disagreements. Table~\ref{tab_stereotypes}-B gives example interpretations of topic \ref{top_acl_2} at different LoA.

\citet{https://doi.org/10.1111/cogs.12189} argue that people are biased to make spontaneous, implicit, and inductive generalizations to certain categories from information given sparse evidence about categories. Such generalizations can be seen in sciences~\cite{https://doi.org/10.1111/cogs.13188} and in political communications~\cite{doi:10.1073/pnas.2309361120}.
Users AI-2, AI-4, and AI-9's interpretation of topic \ref{top_acl_1} in Table~\ref{tab_stereotypes}-C show generalizations of the topic based on the role played by different words in the inductive inference(s) while interpreting the topic.

{\subsection{Gestalt principles}\label{sec_gestalt}}
\begin{table*}[!t]\footnotesize
	\begin{tabular}{P{1.6cm}P{2.6cm}P{10.5cm}}
		\toprule
		\textbf{ID} & \textbf{Topic label} & \textbf{Rationale} \\ \toprule
		\multicolumn{3}{p{14cm}}{\textbf{\textit{A. Gestalt principle: Closure}}} \\ \midrule 
		{SI-11 for \textbf{\ref{top_senate_5}}} & {US and Ukraine conflict with russia and China} & [...] My guess is this topic is about the Ukrain joining Nato struggle and conflicts with Russia[...]\\ \midrule
		{DI-4 for \textbf{\ref{top_design_4}}} & {Concurrency and Processes Management} & For me the words:[...] are related on how different processes or units of execution access the same resources avaliable. "transaction", "update", "block", and "locked" -> are words for me that focus more on the integrity of the data[...]\\ \midrule\midrule
		\multicolumn{3}{p{14cm}}{\textbf{\textit{B. Gestalt principle: Figure and Ground for Topic~\ref{top_senate_4}}}} \\ \midrule 
		SI-4 & American Human Rights & There are words having to do   with different categories of people (women, men, \textbf{black}), words having to do   with rights (equality, amendment, fair, justice,[...] \\ \midrule
		SI-5 & Women’s rights & Things like equality, \textbf{Black},   women, rights, discrimination all point to a speech about Women's rights [...] \\ \midrule
		SI-9 & Equal Rights & After looking at the first few   words "women","rights","act","equal"   \textbf{all of these related towards the movement for equality between women and men} [...] \\ \bottomrule
	\end{tabular}
	\caption{Topic interpretations showing effects of Gestalt principles. Table~\ref{tab_gestalts_emergence_app} provides full texts of the interpretations.}
	\label{tab_gestalts_emergence}
\end{table*}

Table~\ref{tab_gestalts_emergence}-A, shows a few topics in which interpreters use words or concepts that are not mentioned in the respective $T_W$s. SI-11 mentions ``Ukraine's NATO struggle'' even though the word NATO is not mentioned in Topic \ref{top_senate_5}, similarly DI-4 mentions ``concurrency'' while interpreting Topic \ref{top_design_4}.

Such behavior of interpreters to go beyond $T_W$ can be understood through the lens of ``Reader-response theory'' by~\cite{iser1978act}. He argues that while reading a text, the reader strives for consistency and coherence by filling the ``gaps'' in the text based on their imagination and environment. Filling the gaps gives the \textit{gestalt} of the text.

Figure/ground is another important gestalt principle~\cite{koffka1999principles}. In the perceptual decision, a person focuses on certain objects (\emph{figure}) and considers other objects (\emph{ground}) relatively less important. In case of ambiguity, the same objects can be seen as either \emph{figure} or \textit{ground}. Table~\ref{tab_gestalts_emergence}-B provides examples of this principle in topic interpretations. While interpreting Topic \ref{top_senate_4}, interpreter SI-4 considers the word ``black'' as a \textit{figure}, SI-5 also considers it as a \textit{figure} but subsumes ``black identity'' with ``women identity'', SI-9 considers ``black'' as a ground. We argue that the role of a word in interpreting a topic depends on whether it is considered as a figure or ground.

\begin{figure*}
	\centering
	\includegraphics{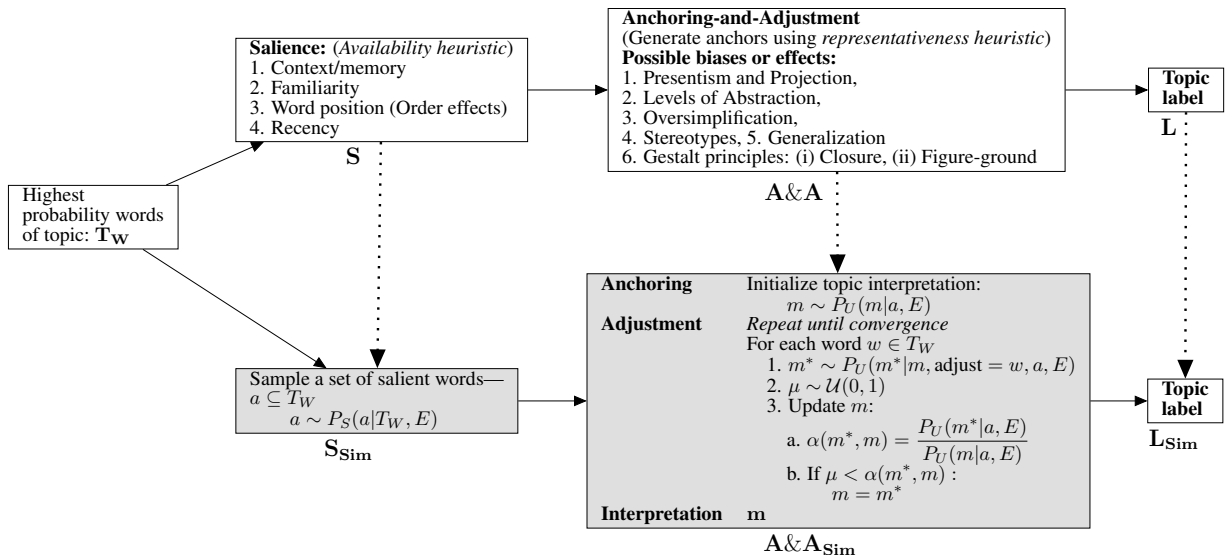}
	\caption{A theory of topic interpretation using the Anchoring-and-Adjustment heuristic along with its computational framework (shaded boxes).}
	\label{fig:theory}
\end{figure*}

\begin{table*}[!t]\footnotesize
	\begin{tabular}{P{0.8cm}P{2.cm}P{11.85cm}}
		\toprule
		\multicolumn{3}{p{15.5cm}}{\textit{\textbf{\ref{top_senate_3}}: freedom religious american history rights government united states americans human nation world america religion cuba liberty war state free faith nations democracy society political cuban}} \\ \midrule
		\textbf{ID} & \textbf{User's Label} & \textbf{Rationale} \\ \midrule 
		{SI-12} &	{Religious Freedom in American Government} &	{
			\textbf{Since "religious" and "freedom" are the top two words}, I would guess that this topic is more specifically referring to religious freedom in American government, but I'm not sure -- I would need to see documents to confirm this. \textbf{Also, it isn't clear to me whether "cuba" and "cuban" belong in this topic.}} \\ \midrule
		{SI-13} &	{Spanish-American War} &	{Words like Cuban, America, war, and freedom indicated to me that the topic is about the Spanish-American War, which primarily revolved around Cuban independence. \textbf{The words related to religion/faith threw me off the most, which led me to lowering my confidence-related scores.}[...]} \\ \bottomrule 
	\end{tabular}
	\caption{Examples of topic interpretations based on the anchoring-and-adjustment heuristic. Table~\ref{tab_aa_examples_app} provides full texts of the interpretations.}
	\label{tab_aa_examples}
\end{table*}

\section{A Theory of Topic Interpretation}\label{sec_theory_ti}
Our theory connecting the themes from  Section~\ref{sec_themes} is based on \textit{the anchoring effect} ($A\&A$), the product of an \textit{anchoring-and-adjustment} ($A\&A$) heuristic~\cite{Tversky_Kahneman_Judgment}. While making judgments under uncertainty, a person \textit{anchors} on information that comes to mind and \textit{adjusts} until a plausible estimate is reached~\cite{Epley_Gilovich_2006}. To understand $A\&A$, consider the question inspired by~\cite{doi:https://doi.org/10.1002/9780470752937.ch12}: ``When did George Washington step down as President of the United States?'' Suppose a person does not know the exact answer. In that case, they may anchor on the fact that the US declared its independence in 1776, and then adjust it based on (i) Washington was the first president; (ii) he must have been elected shortly after 1776; and (iii) he must have stepped down either four or eight years after the election\footnote{\url{https://constitution.congress.gov/browse/essay/artII-S1-C1-9/ALDE_00013597/} Last accessed: 21-July-2025}.

While interpreting a topic, a user (i) generates a set of anchor concepts/categories; (ii) interprets non-anchor words contextually but compatibly with the anchors; (iii) assimilates these interpretations to arrive at an overall interpretation and its encapsulation---a label. 

The two heuristics that play a vital role in anchor generation are: \textit{availability} and \textit{representativeness}~\cite{Tversky_Kahneman_Judgment}. While employing availability heuristics, a user identifies certain words as important based on \textit{ease of retrieval} from memory considering their vividness, associated emotions, exposure, familiarity, recency, etc. The user then uses the representativeness heuristic to identify candidate categories/concepts as anchors based on their similarity with the important words identified in the previous step. The user then makes inductive inferences to adjust the anchors w.r.t. the remaining words and the context.

As the anchors are self-generated they necessarily vary across the users depending on their priming, positionality, and environment. Hence, anchor-adjustments lead to multiple interpretations $I_n$ in the Triangle of Reference (Fig. \ref{fig:triangle_reference}). Such anchoring is similar to self-anchoring in conversations~\cite{Keysar_Barr_2002}, where speakers in case of uncertainty tend to anchor on their egocentric perspective and attempt to adjust to the perspective of others. Often these adjustments are insufficient leading to miscommunication.

Figure~\ref{fig:theory} summarizes our theory of topic interpretations:  frequency, familiarity, and word position influence anchor generation (relation between $T_W$ and $U_i$ in Figure~\ref{fig:triangle_reference}), and effects such as presentism, levels of abstraction, Gestalt principles, and stereotypes influence the subsequent adjustment process (relation between $U_i$ and $I_i$ in Figure~\ref{fig:triangle_reference}). The differences in anchors lead to differences in interpretations, biased towards the anchors.% As per the theory interpetability is not a linear combination of similarity between pair of words. 

We illustrate this theory with examples in Table~\ref{tab_aa_examples}. For topic \ref{top_senate_3}, user SI-12 considers words \texttt{\textit{religious}} and \texttt{\textit{freedom}} as anchors as they are the top two words. They generate \texttt{\textit{American government and politics}} as another anchor based on the words, \texttt{\textit{government, united, states, american}}, etc. Based on these anchors they arrive at the label: \texttt{\textit{Religious Freedom in American Government}}. Importantly, in the adjustment process, they explicitly consider words \texttt{\textit{cuba}} and \texttt{\textit{cuban}} as irrelevant.

For the same topic, user SI-13 considers \texttt{\textit{cuba}} and \texttt{\textit{cuban}} as the anchors, even though they are at lower ranks. They generate \texttt{\textit{Spanish}} and \texttt{\textit{independence}} as other anchors, even though they are not present in $T_W$. Based on other anchors, \texttt{\textit{american, war}} and \texttt{\textit{freedom}} they arrive at the label \texttt{\textit{Spanish-American War}}. In contrast with SI-12, words \texttt{\textit{religion}} and \texttt{\textit{faith}} do not play any role in their adjustment process. In the adjustment process, SI-12 adjusts the word \texttt{\textit{freedom}} towards ``\textit{freedom of religion}'', while SI-13 adjusts the same word towards ``\textit{war of Cuba with Spain}''. Due to the differences in their anchors and further adjustments SI-12 and SI-13 arrive at different interpretations.

\subsection{Topic Interpretation as Approximate Bayesian Inference}\label{cframework}
Several authors, such as~\cite{CHATER2006335,Edward_one_and_done_2014,griffiths2024bayesian} argue that human cognition is parallel to Bayesian inference. As humans are boundedly (or ecologically) rational, the inference is approximate~\cite{Lieder_Griffiths_2020}. \citet{Lieder_thesis} assumes a rational mind that makes effective use of their resources and applies Bayesian decision theory to simulate ``availability'' and ``anchoring-and-adjustment'' heuristics. 

We propose a computational framework for our theory of topic interpretation based on the resource-rational framework of human cognition~\cite{Lieder_thesis}. According to our framework, a user identifies a set of words $a$ as the set of salient words in $T_W$ by sampling from their user-specific distribution---$P_S$ conditioned on their ecology $E$ consisting of their decision context, memories, and experiences. The user then generates an ``anchor interpretation'' ($m$) as the initial interpretation by sampling from $P_U$ conditioned on the $a$ and $E$, where $ P_U$ is a probability distribution over all possible interpretations of the topic. The user then adjusts the anchor based on other words. We model \textit{adjustment} as a Markov chain that starts with the ``anchor interpretation''. At each step, an adjustment ($m^*$) to the current interpretation ($m$) conditioned on word $w$ from $T_W$ is proposed: $m^* \sim P_U(m^*|m, \text{adjust} = w, a, E)$, and it is accepted according to the Metropolis–Hastings algorithm~\cite{Robert_and_Casella_2009}. The adjustments are made until the current interpretation converges. Shaded blocks in Figure~\ref{fig:theory} summarizes the computational framework.

From a practical standpoint, a knowledge source such as Wikipedia or a domain-specific knowledge graph (KG) can be used to simulate our theory. The basic idea is to assume that a topic can be interpreted as a Wikipedia article (or a node in the KG), and its title as the label of the topic. Let $P_U(W_k|T_W)$ be the probability that user $U$ will select the Wikipedia article $W_k$ as an interpretation of topic $T_W$. A user based on a set of salient words ($a$), identifies a Wikipedia article as an anchor interpretation and then adjusts the same using the process described in the $A\&A_{Sim}$ block in Figure~\ref{fig:theory} considering different types of relations (or edges) between Wikipedia articles (or nodes in the KG).

Multiple sets of salient words ($\mathcal{A} = \{a_i\}$) can be sampled from $P_S$ specific to the knowledge source and then each $a_i \in \mathcal{A}$ can be used to generate multiple topic interpretations---$\mathcal{M}$, such that each $m \in \mathcal{M}$ corresponds to a Wikipedia article (or a node) as per our framework described above.

The interpretations in the form of Wikipedia articles (or nodes in a KG) in $\mathcal{M}$ can be used to estimate the quality of topics and therefore compare different topic models: (i) (dis)similarity of the Wikipedia articles in $\mathcal{M}$; (ii) how generic or specific are they; (iii) what would be the effect incorporating word order in identification of salient words and the $A\&A$ process on $\mathcal{M}$; (iv) to study the \textit{presentism} or \textit{projection} effects discussed in Section~\ref{sec_presentism}, older or newer Wikipedia dumps can be used to generate $\mathcal{M}$. 

Depending on the goals of content analysis, \textit{consistency}, \textit{diversity} (a spectrum of possible interpretations), or \textit{divergence} (clusters of similar interpretations)~\cite{DBLP:conf/chi/ZadeDCGA18} among $\mathcal{M}$ can serve as an indicator of the utility of a topic. \textit{Consistency} may be required in the content analysis of laws and regulations or healthcare-related documents, while \textit{diversity} of different perspectives may play a role in the analysis of literature (e.g., poetry) or humanities.

The proposed framework and its utility in evaluation need to be tested rigorously, specifically to arrive at the subjective probability distributions--- $P_S$ and $P_U$, specific to a user. We want to explore the same in the future.

\section{Discussion}\label{sec_discussion}
\subsection{Construct Validity of Coherence Ratings}\label{sec_disc_coherence_ratings}
In Section~\ref{sec_icc}, we observed significant variations at both the user level and the construct level.
The constructs~\ref{s_k2}, \ref{s_c3}, \ref{s_int_exm}, and \ref{s_int_order} can be seen as sources of variations. Moreover, some constructs are significantly correlated. These insights can be used to develop an evaluation framework for topic models. These variations are not captured by the LLM-based coherence ratings ($R_{w/o}^K$ and $R_{w/}^K$) as there is little variation among these scores for both $K=3$ and $K=5$.

We argue that the scores assigned by the LLM are its \textit{revealed preferences} rather than its \textit{normative preferences}. Such preferences are sensitive to \textit{framing} (especially the choices or options provided to the participant)~\cite{framing_Tversky_Kahneman,goldin_revealed_preference_2020,BLOEM2024111686}. A LLM (or even humans) may observe variations within and across topics in reality but these variations do not emerge out (e.g., \ref{top_acl_1} vs. \ref{top_acl_4}) because of the constraints imposed by the options and choices provided to them (in this case $K$). This is a major limitation of the rating-based assessment of topic interpretations.

Our quantitative and qualitative analyses demonstrate that topic interpretation is multidimensional, and a single score or rating has limited validity in assessing topic quality. Many fields of health care (especially Psychology~\cite{psychiatryDSM5Online}), the humanities, and social sciences~\cite{pewresearchDesignedScale} measure latent (and often abstract) social-psychological constructs using multiple items or indicators~\cite{Netemeyer2003}. Carefully designed constructs, items, and their framing may lead to objective testing of a theory. \citet[Figure 1.1]{Netemeyer2003} propose a four-step process to develop a scale for the evaluation of a socio-psychological phenomenon: (i) defining constructs, (ii) generating measurement items, (iii) designing and conducting studies, (iv) finalizing the scale. A similar process with the constructs and items proposed in this paper can be utilized to develop a scale of topic interpretation.

\subsection{Rationality: Axiomatic vs Ecological}
The underlying axioms in automatic coherence measures can be described as: 
(i) a topic's interpretation depends only on its $T_W$, (ii) the interpretability of a topic is proportional to its coherence, (iii) the coherence is proportional to \textbf{a linear combination} of semantic relatedness between every pair of words in $T_W$, (iv) all the users use the same semantic relatedness measure all the time.

Axiomatic rationality assumes a rational agent \textit{should} conform to abstract axioms for profit and their violations will result in loss. It ignores how an agent \textit{will} act in practice, especially when there is uncertainty. Often, humans rely on heuristics that may violate axioms of logic and probability~\cite{Tversky_Kahneman_Judgment}.

Section~\ref{sec_themes} discusses several examples showing how the axioms are violated in topic interpretations. Axiomatic rationality further assumes \textit{the invariance of the measurement}~\cite{https://doi.org/10.1348/000712607X251243} which is valid in laboratory experiments in physical sciences but not in tasks with uncertainty.

\citet{Sen1990} argues that a key goal of using models of rational behavior is to explain and predict actual behavior by (i) characterizing rational behavior, (ii) basing actual behavior on rational behavior. These axioms do not characterize rational and actual behavior. Moreover, $C_{stat}$ and WI are \textit{system-centric}~\cite{10.1145/3556536} and not \textit{user-centric}. Hence, topic interpretability metrics are needed based on ecological rationality, assuming rational behavior as a function of the mind and its environment~\cite{Gigerenzer2021}. One approach to characterize actual behavior is to have \textit{user-models}---a computational representation of how a user will interpret a topic~\cite{aloteibi_2020}.

\subsection{User Models}\label{sec_disc_um}
Topic models make assumptions about \textit{generation} of documents and not on \textit{interpretation} of topics. \citet[Chapter 2]{Krippendorff2019} propose a conceptual framework for QCA that ``includes the researcher, the knowledge he or she needs to bring to it, and the criteria by which a content analysis can be justified'' ($U_i$ in Figure~\ref{fig:triangle_reference}). This framework is not modeled in the underlying assumptions of a topic model and more importantly, its evaluation metrics.
We hypothesize user models can reduce the \textit{generation-interpretation} gap by considering users as an internal entity in topic interpretation, rather than an external one~\cite{aloteibi_2020}. Such models will (i) predict and explain how a user will interpret a topic and (ii) facilitate simulations of the behavior of users across domains, goals, and expertise, which in turn help in designing evaluation metrics, user interfaces, and reproducible experiments~\cite{balog2023user}.

\subsection{Implications} Our key observations from our quantitative ($\S$\ref{sec_quant_analysis}) and qualitative analysis ($\S$\ref{sec_theory_ti}) are: (i) A topic's interpretations in reality fall on a continuum of abstractions,  (ii) A statistically coherent topic can nevertheless have quite different interpretations, (iii) An interpretable topic can be incoherent, (iv) (In)coherence of a topic can be multifaceted. In sum, the interpretation of topics can be viewed as decision-making under uncertainty; hence, interpretations are susceptible to cognitive biases of interpreters~\cite{Gigerenzer2021}.

Coherence and interpretability are two different aspects of a topic and ``one'' quantitative score or rating, either by statistical metrics or human, is insufficient to decide its utility for QCA. Variability in user assessment for both within and across topics reflects a lack of predictability of topic quality. $C_{stat}$ or LLM-based assessment of topics does not capture such variability. Moreover, the end results of topic modeling based studies should be assessed and contested on \textit{reflexivity} and \textit{positionality} of model assumptions and humans or LLM-based interpreters.

\subsection{Topic Cards}   
Conclusions drawn from qualitative research are well-known to be subject to the researcher’s position, subjectivity, and reflexivity~\cite{Krippendorff2019}. To capture subjectivity, and to compare the capabilities of topic models, we propose \textit{Topic Cards}, a new framework similar to Data Cards~\cite{10.1145/3531146.3533231} and Model Cards~\cite{10.1145/3287560.3287596}. A Topic Card should have easy-to-understand explanations and rationales of \textit{who} and \textit{how} topics were interpreted and \textit{what} are the implications of the interpretations. Topic cards will enhance reporting of reflexivity and the subsequent credibility of QCA.

A topic card should at least contain: (i) The task, purpose, goals, or research questions intended to address or achieve using topics, (ii) User positionality, (iii) Rationale behind topic interpretation and label for all the users, (iv) Utility of the topic, (v) Details of prompts if LLMs are used while interpreting topics, and (vi) How sensitive are the results with the interpretations. Appendix~\ref{app_example_topic_card} provides an example of a Topic Card.

\section{Conclusions}
To understand how users interpret topics, we proposed several constructs of topic quality. We asked users (i) to evaluate them on topics from three datasets, (ii) provide rationales for their evaluations. Quantitative analysis of their evaluations shows that users disagree while interpreting topics. Our reflexive thematic analysis of the rationales of users shows users employ heuristics such as Availability and Representativeness and violate axioms of rationality---rules of logic and probability. The thematic analysis also reveals that coherence and interpretability are different and multidimensional constructs that are not captured by state-of-the-art evaluation metrics. The assumptions of interpretability in these metrics are system-centric, and not user-centric. Our analyses show that a rating-based evaluation of topic quality has limited construct validity. Our findings also question the rationality and invariance assumptions behind the state-of-the-art coherence metrics. 

We propose a theory of topic interpretation based on the \textit{anchoring-and-adjustment heuristic} that encompasses several heuristics from psychology and cognitive science. We propose a computational framework to simulate our theory, assuming topic interpretation as an approximate Bayesian inference. The proposed constructs and framework can be utilized to define a new evaluation framework for topic models.

We argue that there is a need for evaluation metrics based on ecological rationality. We propose \textit{Topic Cards} to facilitate reflexivity and validation of the inferences derived from topic interpretations.

\section{Positionality Statement for the Study Authors}\label{sec_positionality}
Swapnil (he/him) is a citizen of India who received his PhD in Computer Science and Engineering from IIT Madras. He is an Assistant Professor at IIT Palakkad. His primary research interests are in natural language processing (NLP) and evaluation metrics in AI. In his previous works, he explored approaches for reducing knowledge acquisition overhead in NLP tasks. He was employed at the University of Victoria, while doing this work.

Ze Shi Li is an assistant professor at the University of Oklahoma. He received his PhD at the University of Victoria. He is an incoming Assistant Professor at the University of Oklahoma. His research focuses on AI for software engineering and human-centered AI.

Vivienne Zeng (she/her) is a Chinese citizen and a master's student in Computer Science at the University of Victoria. Her research interests include AI-assisted software engineering and large language models, with a focus on their application in scientific software development. She contributed to this work as part of her graduate research under the supervision of Dr. Neil Ernst at University of Victoria.

Ahmed (he/him) is a Bangladeshi citizen who received his Master’s degree in Computer Science from the University of Victoria. His academic and professional interests lie at the intersection of software engineering and natural language processing (NLP). He was a Master’s student at University of Victoria while conducting this work.

Luiz Pedro Franciscatto Guerra (he/him) is a citizen of Brazil who received his Bachelor's degree in Software Engineering from Pontifícia Universidade do Rio Grande do Sul. His academic interests lie at the intersection of Software Engineering, Software Architecture, and Large Language Models. He's a computer science PhD student at University of Victoria while conducting this work.

Neil Ernst (he/him) is a Canadian citizen who received his PhD from the University of Toronto. He is Associate Professor of Computer Science at the University of Victoria where he works on the intersection of AI and software engineering.

\section*{Acknowledgments}
We thank Jordan Boyd-Graber, Sutanu Chakraborti, Caitlin Doogan, Alexander Hoyle, Zongxia Li, Jia Peng Lim, Alessandra Maciel Paz Milani, David Mimno, Deepak P, Arty Starr, Sneha Swami, Keyon Vafa for their inputs on user studies and qualitative analysis. We also thank the action editor and the anonymous reviewers for their valuable feedback on prior versions of this work.
\bibliography{tacl2021}

\begin{thebibliography}{102}
\expandafter\ifx\csname natexlab\endcsname\relax\def\natexlab#1{#1}\fi

\bibitem[{Aloteibi(2020)}]{aloteibi_2020}
Saad Aloteibi. 2020.
\newblock \href {https://doi.org/10.17863/CAM.77137} {\emph{A user-centred
  approach to information retrieval}}.
\newblock Ph.D. thesis, University of Cambridge Repository.

\bibitem[{{American Psychiatric Association}(2025)}]{psychiatryDSM5Online}
{American Psychiatric Association}. 2025.
\newblock \href
  {https://www.psychiatry.org/psychiatrists/practice/dsm/educational-resources/dsm-5-assessment-measures}
  {{D}{S}{M}-5 {O}nline {A}ssessment {M}easures}.
\newblock
  \url{https://www.psychiatry.org/psychiatrists/practice/dsm/educational-resources/dsm-5-assessment-measures}.
\newblock [Accessed 18-06-2025].

\bibitem[{Anderson and Schooler(1991)}]{Anderson_Schooler}
John~R. Anderson and Lael~J. Schooler. 1991.
\newblock \href {https://doi.org/10.1111/j.1467-9280.1991.tb00174.x}
  {{Reflections of the Environment in Memory}}.
\newblock \emph{Psychological Science}, 2(6):396--408.

\bibitem[{Aroyo and Welty(2015)}]{DBLP:journals/aim/AroyoW15}
Lora Aroyo and Chris Welty. 2015.
\newblock \href {https://doi.org/10.1609/aimag.v36i1.2564} {{Truth Is a Lie:
  Crowd Truth and the Seven Myths of Human Annotation}}.
\newblock \emph{{AI} Mag.}, 36(1):15--24.

\bibitem[{Azzopardi(2021)}]{conf/chiir/Azzopardi21}
Leif Azzopardi. 2021.
\newblock \href {https://doi.org/10.1145/3406522.3446023} {Cognitive biases in
  search: {A} review and reflection of cognitive biases in information
  retrieval}.
\newblock In \emph{{CHIIR} '21: {ACM} {SIGIR} Conference on Human Information
  Interaction and Retrieval, Canberra, ACT, Australia, March 14-19, 2021},
  pages 27--37. {ACM}.

\bibitem[{Balog and Zhai(2023)}]{balog2023user}
Krisztian Balog and ChengXiang Zhai. 2023.
\newblock \href {https://doi.org/10.48550/ARXIV.2306.08550} {{User Simulation
  for Evaluating Information Access Systems}}.
\newblock \emph{CoRR}, abs/2306.08550v1.

\bibitem[{Baltes et~al.(2019)Baltes, Treude, and
  Diehl}]{DBLP:conf/msr/BaltesT019}
Sebastian Baltes, Christoph Treude, and Stephan Diehl. 2019.
\newblock \href {https://doi.org/10.1109/MSR.2019.00038} {Sotorrent: studying
  the origin, evolution, and usage of stack overflow code snippets}.
\newblock In \emph{Proceedings of the 16th International Conference on Mining
  Software Repositories, {MSR} 2019, 26-27 May 2019, Montreal, Canada}, pages
  191--194. {IEEE} / {ACM}.

\bibitem[{Belz and Kow(2010)}]{DBLP:conf/inlg/BelzK10}
Anja Belz and Eric Kow. 2010.
\newblock \href {https://aclanthology.org/W10-4201/} {Comparing rating scales
  and preference judgements in language evaluation}.
\newblock In \emph{{INLG} 2010 - Proceedings of the Sixth International Natural
  Language Generation Conference, July 7-9, 2010, Trim, Co. Meath, Ireland}.
  The Association for Computer Linguistics.

\bibitem[{Berger(2015)}]{doi:10.1177/1468794112468475}
Roni Berger. 2015.
\newblock \href {https://doi.org/10.1177/1468794112468475} {{Now I see it, now
  I don't: researcher's position and reflexivity in qualitative research}}.
\newblock \emph{Qualitative Research}, 15(2):219--234.

\bibitem[{Bernell(2012)}]{Bernell2012}
David Bernell. 2012.
\newblock \href {https://doi.org/10.4324/9780203829264} {\emph{Constructing US
  Foreign Policy The Curious Case of Cuba}}.
\newblock Routledge.

\bibitem[{Beshears et~al.(2008)Beshears, Choi, Laibson, and
  Madrian}]{BESHEARS20081787}
John Beshears, James~J. Choi, David Laibson, and Brigitte~C. Madrian. 2008.
\newblock \href {https://doi.org/https://doi.org/10.1016/j.jpubeco.2008.04.010}
  {How are preferences revealed?}
\newblock \emph{Journal of Public Economics}, 92(8):1787--1794.
\newblock Special Issue: Happiness and Public Economics.

\bibitem[{Blair and Kimbrough(2002)}]{BLAIR2002363}
David~C Blair and Steven~O Kimbrough. 2002.
\newblock \href {https://doi.org/https://doi.org/10.1016/S0306-4573(01)00027-9}
  {Exemplary documents: a foundation for information retrieval design}.
\newblock \emph{Information Processing \& Management}, 38(3):363--379.

\bibitem[{Bloem and Rahman(2024)}]{BLOEM2024111686}
Jeffrey~R. Bloem and Khandker~Wahedur Rahman. 2024.
\newblock \href {https://doi.org/https://doi.org/10.1016/j.econlet.2024.111686}
  {{What I say depends on how you ask: Experimental evidence of the effect of
  framing on the measurement of attitudes}}.
\newblock \emph{Economics Letters}, 238:111686.

\bibitem[{Boswijk(2022)}]{Boswijk_salience}
{Vincent H.} Boswijk. 2022.
\newblock \href {https://doi.org/10.33612/diss.203005779} {\emph{The salient
  elephant in the room: exploring the concept of linguistic salience}}.
\newblock Ph.D. thesis, University of Groningen.

\bibitem[{Braun and Clarke(2012)}]{Braun_Clarke_2012}
Virginia Braun and Victoria Clarke. 2012.
\newblock \href {https://doi.org/10.1037/13620-004} {\emph{Thematic analysis.}}
\newblock APA handbooks in psychology®. American Psychological Association,
  Washington, DC, US.

\bibitem[{Braun and Clarke(2021)}]{Braun_Clarke_2021}
Virginia Braun and Victoria Clarke. 2021.
\newblock \href {https://doi.org/https://doi.org/10.1002/capr.12360} {{Can I
  use TA? Should I use TA? Should I not use TA? Comparing reflexive thematic
  analysis and other pattern-based qualitative analytic approaches}}.
\newblock \emph{Counselling and Psychotherapy Research}, 21(1):37--47.

\bibitem[{Bubeck et~al.(2023)Bubeck, Chandrasekaran, Eldan, Gehrke, Horvitz,
  Kamar, Lee, Lee, Li, Lundberg, Nori, Palangi, Ribeiro, and
  Zhang}]{bubeck2023sparks}
Sébastien Bubeck, Varun Chandrasekaran, Ronen Eldan, Johannes Gehrke, Eric
  Horvitz, Ece Kamar, Peter Lee, Yin~Tat Lee, Yuanzhi Li, Scott Lundberg,
  Harsha Nori, Hamid Palangi, Marco~Tulio Ribeiro, and Yi~Zhang. 2023.
\newblock \href
  {https://www.microsoft.com/en-us/research/publication/sparks-of-artificial-general-intelligence-early-experiments-with-gpt-4/}
  {{Sparks of Artificial General Intelligence: Early experiments with GPT-4}}.

\bibitem[{Chang et~al.(2009)Chang, Boyd{-}Graber, Gerrish, Wang, and
  Blei}]{DBLP:conf/nips/ChangBGWB09}
Jonathan~D. Chang, Jordan~L. Boyd{-}Graber, Sean Gerrish, Chong Wang, and
  David~M. Blei. 2009.
\newblock \href
  {https://proceedings.neurips.cc/paper/2009/hash/f92586a25bb3145facd64ab20fd554ff-Abstract.html}
  {Reading tea leaves: How humans interpret topic models}.
\newblock In \emph{Advances in Neural Information Processing Systems 22: 23rd
  Annual Conference on Neural Information Processing Systems 2009. Proceedings
  of a meeting held 7-10 December 2009, Vancouver, British Columbia, Canada},
  pages 288--296. Curran Associates, Inc.

\bibitem[{Chater and Manning(2006)}]{CHATER2006335}
Nick Chater and Christopher~D. Manning. 2006.
\newblock \href {https://doi.org/https://doi.org/10.1016/j.tics.2006.05.006}
  {Probabilistic models of language processing and acquisition}.
\newblock \emph{Trends in Cognitive Sciences}, 10(7):335--344.
\newblock Special issue: Probabilistic models of cognition.

\bibitem[{Clancy and Silver(2022)}]{pewresearchDesignedScale}
Laura Clancy and Laura Silver. 2022.
\newblock {H}ow we designed a scale to measure {A}mericans’ knowledge of
  international affairs.
\newblock
  \url{https://www.pewresearch.org/decoded/2022/05/25/how-we-designed-a-scale-to-measure-americans-knowledge-of-international-affairs/}.
\newblock [Accessed 18-06-2025].

\bibitem[{Collier et~al.(2006)Collier, Hidalgo, and
  Maciuceanu}]{Collier_contested_concepts}
David Collier, Fernando~Daniel Hidalgo, and Andra~Olivia Maciuceanu. 2006.
\newblock \href {https://doi.org/10.1080/13569310600923782} {Essentially
  contested concepts: Debates and applications}.
\newblock \emph{Journal of Political Ideologies}, 11(3):211--246.

\bibitem[{DiMaggio et~al.(2013)DiMaggio, Nag, and Blei}]{DIMAGGIO2013570}
Paul DiMaggio, Manish Nag, and David Blei. 2013.
\newblock \href {https://doi.org/https://doi.org/10.1016/j.poetic.2013.08.004}
  {Exploiting affinities between topic modeling and the sociological
  perspective on culture: Application to newspaper coverage of u.s. government
  arts funding}.
\newblock \emph{Poetics}, 41(6):570--606.
\newblock Topic Models and the Cultural Sciences.

\bibitem[{Dionne and Seay(2016)}]{Dionne20208}
Kim~Yi Dionne and Laura Seay. 2016.
\newblock {American} {Perceptions} of {Africa} during an {Ebola} {Outbreak}.
\newblock In Nicholas~G. Evans, Tara~C. Smith, and Maimuna~S. Majumder,
  editors, \emph{Ebola's {Message}}. {MIT Press}.

\bibitem[{Doogan and Buntine(2021)}]{doogan-buntine-2021-topic}
Caitlin Doogan and Wray Buntine. 2021.
\newblock \href {https://doi.org/10.18653/v1/2021.naacl-main.300} {Topic model
  or topic twaddle? re-evaluating semantic interpretability measures}.
\newblock In \emph{Proceedings of the 2021 Conference of the North American
  Chapter of the Association for Computational Linguistics: Human Language
  Technologies}, pages 3824--3848, Online. Association for Computational
  Linguistics.

\bibitem[{Dupret and Piwowarski(2013)}]{DUPRET201349}
Georges Dupret and Benjamin Piwowarski. 2013.
\newblock \href {https://doi.org/https://doi.org/10.1016/j.jda.2012.10.002}
  {{Model Based Comparison of Discounted Cumulative Gain and Average
  Precision}}.
\newblock \emph{Journal of Discrete Algorithms}, 18:49--62.
\newblock Selected papers from the 18th International Symposium on String
  Processing and Information Retrieval (SPIRE 2011).

\bibitem[{Egami et~al.(2022)Egami, Fong, Grimmer, Roberts, and
  Stewart}]{Egami_Fong_Grimmer_Roberts_Stewart}
Naoki Egami, Christian~J. Fong, Justin Grimmer, Margaret~E. Roberts, and
  Brandon~M. Stewart. 2022.
\newblock \href {https://doi.org/10.1126/sciadv.abg2652} {How to make causal
  inferences using texts}.
\newblock \emph{Science Advances}, 8(42):eabg2652.

\bibitem[{Enkvist(1990)}]{Enkvist1990May01}
Nils~Erik Enkvist. 1990.
\newblock \href
  {http://search.proquest.com.ezproxy.library.uvic.ca/scholarly-journals/discourse-comprehension-text-strategies-style/docview/1311106132/se-2?accountid=14846}
  {{Discourse Comprehension, Text Strategies and Style}}.
\newblock \emph{A.U.M.L.A.: Journal of the Australasian Universities Modern
  Language Association}, 0(73):166.
\newblock Last updated - 2013-02-25.

\bibitem[{Epley(2004)}]{doi:https://doi.org/10.1002/9780470752937.ch12}
Nicholas Epley. 2004.
\newblock \href {https://doi.org/https://doi.org/10.1002/9780470752937.ch12}
  {\emph{A Tale of Tuned Decks? Anchoring as Accessibility and Anchoring as
  Adjustment}}, chapter~12. John Wiley \& Sons, Ltd.

\bibitem[{Epley and Gilovich(2006)}]{Epley_Gilovich_2006}
Nicholas Epley and Thomas Gilovich. 2006.
\newblock \href {https://doi.org/10.1111/j.1467-9280.2006.01704.x} {{The
  Anchoring-and-Adjustment Heuristic: Why the Adjustments Are Insufficient}}.
\newblock \emph{Psychological Science}, 17(4):311--318.
\newblock PMID: 16623688.

\bibitem[{Ethayarajh and Jurafsky(2022)}]{DBLP:conf/emnlp/EthayarajhJ22}
Kawin Ethayarajh and Dan Jurafsky. 2022.
\newblock \href {https://aclanthology.org/2022.emnlp-main.406} {The
  authenticity gap in human evaluation}.
\newblock In \emph{Proceedings of the 2022 Conference on Empirical Methods in
  Natural Language Processing, {EMNLP} 2022, Abu Dhabi, United Arab Emirates,
  December 7-11, 2022}, pages 6056--6070. Association for Computational
  Linguistics.

\bibitem[{Gatt and Belz(2010)}]{DBLP:conf/eacl/GattB10}
Albert Gatt and Anja Belz. 2010.
\newblock \href {https://doi.org/10.1007/978-3-642-15573-4\_14} {Introducing
  shared tasks to {NLG:} the {TUNA} shared task evaluation challenges}.
\newblock In \emph{Empirical Methods in Natural Language Generation:
  Data-oriented Methods and Empirical Evaluation}, volume 5790 of \emph{Lecture
  Notes in Computer Science}, pages 264--293. Springer.

\bibitem[{Gentzkow et~al.(2019)Gentzkow, Shapiro, and
  Taddy}]{stanfordCongressionalRecord}
Matthew Gentzkow, Jesse~M. Shapiro, and Matt Taddy. 2019.
\newblock \href {https://doi.org/https://doi.org/10.3982/ECTA16566} {{Measuring
  Group Differences in High-Dimensional Choices: Method and Application to
  Congressional Speech}}.
\newblock \emph{Econometrica}, 87(4):1307--1340.

\bibitem[{Gigerenzer(2021)}]{Gigerenzer2021}
Gerd Gigerenzer. 2021.
\newblock \href {https://doi.org/10.1007/s11229-019-02296-5} {Axiomatic
  rationality and ecological rationality}.
\newblock \emph{Synthese}, 198(4):3547--3564.

\bibitem[{Givón(1993)}]{jbp_givon}
T.~Givón. 1993.
\newblock \href {https://doi.org/https://doi.org/10.1075/pc.1.2.01giv}
  {Coherence in text, coherence in mind}.
\newblock \emph{Pragmatics \& Cognition}, 1(2):171--227.

\bibitem[{Goldin and Reck(2020)}]{goldin_revealed_preference_2020}
Jacob Goldin and Daniel Reck. 2020.
\newblock \href {https://doi.org/10.1086/706860} {{Revealed-Preference Analysis
  with Framing Effects}}.
\newblock \emph{Journal of Political Economy}, 128(7):2759--2795.

\bibitem[{Griffiths et~al.(2024)Griffiths, Chater, and
  Tenenbaum}]{griffiths2024bayesian}
T.L. Griffiths, N.~Chater, and J.B. Tenenbaum. 2024.
\newblock \emph{Bayesian Models of Cognition: Reverse Engineering the Mind}.
\newblock MIT Press.

\bibitem[{Grimmer et~al.(2021)Grimmer, Roberts, and
  Stewart}]{annurev:/content/journals/10.1146/annurev-polisci-053119-015921}
Justin Grimmer, Margaret~E. Roberts, and Brandon~M. Stewart. 2021.
\newblock \href
  {https://doi.org/https://doi.org/10.1146/annurev-polisci-053119-015921}
  {{Machine Learning for Social Science: An Agnostic Approach}}.
\newblock \emph{Annual Review of Political Science}, 24(Volume 24,
  2021):395--419.

\bibitem[{Hall et~al.(2008)Hall, Jurafsky, and
  Manning}]{DBLP:conf/emnlp/HallJM08}
David Hall, Daniel Jurafsky, and Christopher~D. Manning. 2008.
\newblock \href {https://aclanthology.org/D08-1038/} {Studying the history of
  ideas using topic models}.
\newblock In \emph{2008 Conference on Empirical Methods in Natural Language
  Processing, {EMNLP} 2008, Proceedings of the Conference, 25-27 October 2008,
  Honolulu, Hawaii, USA, {A} meeting of SIGDAT, a Special Interest Group of the
  {ACL}}, pages 363--371. {ACL}.

\bibitem[{Hall(2021)}]{Stuart_Hall_Race}
Stuart Hall. 2021.
\newblock \href {https://doi.org/doi:10.1515/9781478021223-022} {Race, the
  floating signifier: What more is there to say about ``race''?}
\newblock In Paul Gilroy and Ruth~Wilson Gilmore, editors, \emph{Selected
  Writings on Race and Difference}, pages 359--373. Duke University Press, New
  York, USA.

\bibitem[{Hindle et~al.(2015)Hindle, Bird, Zimmermann, and
  Nagappan}]{DBLP:journals/ese/HindleBZN15}
Abram Hindle, Christian Bird, Thomas Zimmermann, and Nachiappan Nagappan. 2015.
\newblock \href {https://doi.org/10.1007/s10664-014-9312-1} {Do topics make
  sense to managers and developers?}
\newblock \emph{Empir. Softw. Eng.}, 20(2):479--515.

\bibitem[{ten Hove et~al.(2024)ten Hove, Jorgensen, and van~der
  Ark}]{2022-94730-001}
Debby ten Hove, Terrence~D. Jorgensen, and L.~Andries van~der Ark. 2024.
\newblock \href {https://doi.org/10.1037/met0000516} {Updated guidelines on
  selecting an intraclass correlation coefficient for interrater reliability,
  with applications to incomplete observational designs.}
\newblock \emph{Psychological Methods}, 29(5):967--979.

\bibitem[{Hoyle et~al.(2021)Hoyle, Goel, Hian{-}Cheong, Peskov, Boyd{-}Graber,
  and Resnik}]{DBLP:conf/nips/HoyleGHPBR21}
Alexander~Miserlis Hoyle, Pranav Goel, Andrew Hian{-}Cheong, Denis Peskov,
  Jordan~L. Boyd{-}Graber, and Philip Resnik. 2021.
\newblock \href
  {https://proceedings.neurips.cc/paper/2021/hash/0f83556a305d789b1d71815e8ea4f4b0-Abstract.html}
  {Is automated topic model evaluation broken? the incoherence of coherence}.
\newblock In \emph{Advances in Neural Information Processing Systems 34: Annual
  Conference on Neural Information Processing Systems 2021, NeurIPS 2021,
  December 6-14, 2021, virtual}, pages 2018--2033.

\bibitem[{Hoyle et~al.(2022)Hoyle, Sarkar, Goel, and
  Resnik}]{DBLP:conf/emnlp/HoyleSGR22}
Alexander~Miserlis Hoyle, Rupak Sarkar, Pranav Goel, and Philip Resnik. 2022.
\newblock \href {https://doi.org/10.18653/v1/2022.findings-emnlp.390} {Are
  neural topic models broken?}
\newblock In \emph{Findings of the Association for Computational Linguistics:
  {EMNLP} 2022, Abu Dhabi, United Arab Emirates, December 7-11, 2022}, pages
  5321--5344. Association for Computational Linguistics.

\bibitem[{Hsieh and Shannon(2005)}]{Hsieh_Shanon_QCA}
Hsiu-Fang Hsieh and Sarah~E. Shannon. 2005.
\newblock \href {https://doi.org/10.1177/1049732305276687} {{Three Approaches
  to Qualitative Content Analysis}}.
\newblock \emph{Qualitative Health Research}, 15(9):1277--1288.
\newblock PMID: 16204405.

\bibitem[{Hunt(2002)}]{hunt2002against}
Lynn Hunt. 2002.
\newblock Against presentism.
\newblock \emph{Perspectives on history}, 40(5):7--9.

\bibitem[{Iser(1978)}]{iser1978act}
Wolfgang Iser. 1978.
\newblock \href {https://books.google.co.in/books?id=eNViAAAAMAAJ} {\emph{The
  Act of Reading: A Theory of Aesthetic Response}}.
\newblock Johns Hopkins paperback. Johns Hopkins University Press.

\bibitem[{Keysar and Barr(2002)}]{Keysar_Barr_2002}
Boaz Keysar and Dale~J. Barr. 2002.
\newblock Self-anchoring in conversation: Why language users do not do what
  they “should”.
\newblock In Thomas Gilovich, Dale Griffin, and Daniel Kahneman, editors,
  \emph{Heuristics and Biases: The Psychology of Intuitive Judgment}, page
  150–166. Cambridge University Press.

\bibitem[{Kingstone et~al.(2008)Kingstone, Smilek, and
  Eastwood}]{https://doi.org/10.1348/000712607X251243}
Alan Kingstone, Daniel Smilek, and John~D. Eastwood. 2008.
\newblock \href {https://doi.org/https://doi.org/10.1348/000712607X251243}
  {{Cognitive Ethology: A new approach for studying human cognition}}.
\newblock \emph{British Journal of Psychology}, 99(3):317--340.

\bibitem[{Kleinberg et~al.(2024)Kleinberg, Ludwig, Mullainathan, and
  Raghavan}]{doi:10.1177/17456916231212138}
Jon Kleinberg, Jens Ludwig, Sendhil Mullainathan, and Manish Raghavan. 2024.
\newblock \href {https://doi.org/10.1177/17456916231212138} {{The Inversion
  Problem: Why Algorithms Should Infer Mental State and Not Just Predict
  Behavior}}.
\newblock \emph{Perspectives on Psychological Science}, 19(5):827--838.
\newblock PMID: 38085919.

\bibitem[{Koffka(1999)}]{koffka1999principles}
K.~Koffka. 1999.
\newblock \href {https://books.google.co.in/books?id=cLnqI3dvi4kC}
  {\emph{Principles of Gestalt Psychology}}.
\newblock Cognitive psychology. Routledge.

\bibitem[{Koo and Li(2016)}]{Koo2016-nz}
Terry~K Koo and Mae~Y Li. 2016.
\newblock A guideline of selecting and reporting intraclass correlation
  coefficients for reliability research.
\newblock \emph{J. Chiropr. Med.}, 15(2):155--163.

\bibitem[{Krippendorff(2019)}]{Krippendorff2019}
Klaus Krippendorff. 2019.
\newblock \href {https://doi.org/10.4135/9781071878781} {\emph{Content
  Analysis: An Introduction to Its Methodology}}, fourth edition.
\newblock SAGE Publications, Inc., Thousand Oaks.

\bibitem[{Kuhn(1962)}]{kuhn_ssr}
Thomas Kuhn. 1962.
\newblock \emph{The Structure of Scientific Revolutions}.
\newblock University of Chicago Press.

\bibitem[{Lakoff and Johnson(2008)}]{lakoff2008metaphors}
G.~Lakoff and M.~Johnson. 2008.
\newblock \href {https://books.google.co.in/books?id=r6nOYYtxzUoC}
  {\emph{Metaphors We Live By}}.
\newblock University of Chicago Press.

\bibitem[{Lau et~al.(2011)Lau, Grieser, Newman, and
  Baldwin}]{DBLP:conf/acl/LauGNB11}
Jey~Han Lau, Karl Grieser, David Newman, and Timothy Baldwin. 2011.
\newblock \href {https://aclanthology.org/P11-1154/} {Automatic labelling of
  topic models}.
\newblock In \emph{The 49th Annual Meeting of the Association for Computational
  Linguistics: Human Language Technologies, Proceedings of the Conference,
  19-24 June, 2011, Portland, Oregon, {USA}}, pages 1536--1545. The Association
  for Computer Linguistics.

\bibitem[{Li et~al.(2024)Li, Mao, Stephens, Goel, Walpole, Dima, Fung, and
  Boyd-Graber}]{Li:Mao:Stephens:Goel:Walpole:Fung:Dima:Boyd-Graber-2024}
Zongxia Li, Andrew Mao, Daniel Stephens, Pranav Goel, Emily Walpole, Alden
  Dima, Juan Fung, and Jordan Boyd-Graber. 2024.
\newblock \href {http://arxiv.org/abs/2401.16348} {Beyond automated evaluation
  metrics: Evaluating topic models on practical social science content analysis
  tasks}.

\bibitem[{Lieder(2018)}]{Lieder_thesis}
Falk Lieder. 2018.
\newblock \emph{{Beyond bounded rationality: Reverse-engineering and enhancing
  human intelligence}}.
\newblock Phd thesis, UC Berkeley.
\newblock Available at \url{https://escholarship.org/uc/item/0mh5z130}.

\bibitem[{Lieder and Griffiths(2020)}]{Lieder_Griffiths_2020}
Falk Lieder and Thomas~L. Griffiths. 2020.
\newblock \href {https://doi.org/10.1017/S0140525X1900061X} {Resource-rational
  analysis: Understanding human cognition as the optimal use of limited
  computational resources}.
\newblock \emph{Behavioral and Brain Sciences}, 43:e1.

\bibitem[{Lim and Lauw(2024)}]{10.1162/coli_a_00518}
Jia~Peng Lim and Hady~W. Lauw. 2024.
\newblock \href {https://doi.org/10.1162/coli_a_00518} {{Aligning Human and
  Computational Coherence Evaluations}}.
\newblock \emph{Computational Linguistics}, pages 1--58.

\bibitem[{Mahadi et~al.(2020)Mahadi, Tongay, and
  Ernst}]{DBLP:conf/wcre/MahadiTE20}
Alvi Mahadi, Karan Tongay, and Neil~A. Ernst. 2020.
\newblock \href {https://doi.org/10.1109/SANER48275.2020.9054792}
  {Cross-dataset design discussion mining}.
\newblock In \emph{27th {IEEE} International Conference on Software Analysis,
  Evolution and Reengineering, {SANER} 2020, London, ON, Canada, February
  18-21, 2020}, pages 149--160. {IEEE}.

\bibitem[{Marani et~al.(2022)Marani, Levine, and
  Baumer}]{DBLP:conf/cikm/MaraniLB22}
Amin~Hosseiny Marani, Joshua Levine, and Eric P.~S. Baumer. 2022.
\newblock \href {https://doi.org/10.1145/3511808.3557410} {One rating to rule
  them all?: Evidence of multidimensionality in human assessment of topic
  labeling quality}.
\newblock In \emph{Proceedings of the 31st {ACM} International Conference on
  Information {\&} Knowledge Management, Atlanta, GA, USA, October 17-21,
  2022}, pages 768--779. {ACM}.

\bibitem[{McCallum(2002)}]{McCallumMALLET}
Andrew~Kachites McCallum. 2002.
\newblock Mallet: A machine learning for language toolkit.
\newblock Http://mallet.cs.umass.edu/dist/mallet-2.0.8.tar.gz.

\bibitem[{Mimno(2023)}]{MimnoArxiv2023_07}
David Mimno. 2023.
\newblock \href {https://mimno.infosci.cornell.edu/arxivcl/030223/} {{Topic
  Modeling on Abstract Submissions to arXiv in the Computing and Language
  section (cs.CL) }}.
\newblock \url{https://mimno.infosci.cornell.edu/arxivcl/030223/}.
\newblock Accessed: 21-July-2025.

\bibitem[{Mitchell et~al.(2019)Mitchell, Wu, Zaldivar, Barnes, Vasserman,
  Hutchinson, Spitzer, Raji, and Gebru}]{10.1145/3287560.3287596}
Margaret Mitchell, Simone Wu, Andrew Zaldivar, Parker Barnes, Lucy Vasserman,
  Ben Hutchinson, Elena Spitzer, Inioluwa~Deborah Raji, and Timnit Gebru. 2019.
\newblock \href {https://doi.org/10.1145/3287560.3287596} {Model cards for
  model reporting}.
\newblock In \emph{Proceedings of the Conference on Fairness, Accountability,
  and Transparency}, FAT* '19, page 220–229, New York, NY, USA. Association
  for Computing Machinery.

\bibitem[{Morewedge et~al.(2023)Morewedge, Mullainathan, Naushan, Sunstein,
  Kleinberg, Raghavan, and Ludwig}]{Morewedge2023}
Carey~K. Morewedge, Sendhil Mullainathan, Haaya~F. Naushan, Cass~R. Sunstein,
  Jon Kleinberg, Manish Raghavan, and Jens~O. Ludwig. 2023.
\newblock \href {https://doi.org/10.1038/s41562-023-01724-4} {Human bias in
  algorithm design}.
\newblock \emph{Nature Human Behaviour}, 7(11):1822--1824.

\bibitem[{Morstatter and Liu(2018)}]{JMLR:v18:17-069}
Fred Morstatter and Huan Liu. 2018.
\newblock \href {http://jmlr.org/papers/v18/17-069.html} {{In Search of
  Coherence and Consensus: Measuring the Interpretability of Statistical
  Topics}}.
\newblock \emph{Journal of Machine Learning Research}, 18(169):1--32.

\bibitem[{Netemeyer et~al.(2003)Netemeyer, Bearden, and Sharma}]{Netemeyer2003}
Richard Netemeyer, William Bearden, and Subhash Sharma. 2003.
\newblock \emph{Scaling Procedures: Issues and Applications}.
\newblock SAGE Publications.

\bibitem[{Newman et~al.(2010)Newman, Lau, Grieser, and
  Baldwin}]{DBLP:conf/naacl/NewmanLGB10}
David Newman, Jey~Han Lau, Karl Grieser, and Timothy Baldwin. 2010.
\newblock \href {https://aclanthology.org/N10-1012/} {Automatic evaluation of
  topic coherence}.
\newblock In \emph{Human Language Technologies: Conference of the North
  American Chapter of the Association of Computational Linguistics,
  Proceedings, June 2-4, 2010, Los Angeles, California, {USA}}, pages 100--108.
  The Association for Computational Linguistics.

\bibitem[{Novoa et~al.(2023)Novoa, Echelbarger, Gelman, and
  Gelman}]{doi:10.1073/pnas.2309361120}
Gustavo Novoa, Margaret Echelbarger, Andrew Gelman, and Susan~A. Gelman. 2023.
\newblock \href {https://doi.org/10.1073/pnas.2309361120} {Generically
  partisan: Polarization in political communication}.
\newblock \emph{Proceedings of the National Academy of Sciences},
  120(47):e2309361120.

\bibitem[{Ogden and Richards(1927)}]{ogden1927meaning}
Charles~Kay Ogden and Ivor~Armstrong Richards. 1927.
\newblock \emph{The Meaning of Meaning: A Study of the Influence of Language
  upon Thought and of the Science of Symbolism}, volume~29.
\newblock K. Paul, Trench, Trubner \& Company, Limited.

\bibitem[{Pei et~al.(2022)Pei, Ananthasubramaniam, Wang, Zhou, Dedeloudis,
  Sargent, and Jurgens}]{DBLP:conf/emnlp/PeiAWZDSJ22}
Jiaxin Pei, Aparna Ananthasubramaniam, Xingyao Wang, Naitian Zhou, Apostolos
  Dedeloudis, Jackson Sargent, and David Jurgens. 2022.
\newblock \href {https://doi.org/10.18653/V1/2022.EMNLP-DEMOS.33} {{POTATO:}
  the portable text annotation tool}.
\newblock In \emph{Proceedings of the The 2022 Conference on Empirical Methods
  in Natural Language Processing, {EMNLP} 2022 - System Demonstrations, Abu
  Dhabi, UAE, December 7-11, 2022}, pages 327--337. Association for
  Computational Linguistics.

\bibitem[{Pereira et~al.(2023)Pereira, Viegas, Gon\c{c}alves, and
  Rocha}]{10.1145/3617023.3617040}
Ant\^{o}nio Pereira, Felipe Viegas, Marcos~Andr\'{e} Gon\c{c}alves, and
  Leonardo Rocha. 2023.
\newblock \href {https://doi.org/10.1145/3617023.3617040} {Evaluating the
  limits of the current evaluation metrics for topic modeling}.
\newblock In \emph{Proceedings of the 29th Brazilian Symposium on Multimedia
  and the Web}, WebMedia '23, page 119–127, New York, NY, USA. Association
  for Computing Machinery.

\bibitem[{Peters et~al.(2022)Peters, Krauss, and
  Braganza}]{https://doi.org/10.1111/cogs.13188}
Uwe Peters, Alexander Krauss, and Oliver Braganza. 2022.
\newblock \href {https://doi.org/https://doi.org/10.1111/cogs.13188}
  {{Generalization Bias in Science}}.
\newblock \emph{Cognitive Science}, 46(9):e13188.

\bibitem[{Popa and Rebedea(2021)}]{popa-rebedea-2021-bart}
Cristian Popa and Traian Rebedea. 2021.
\newblock \href {https://doi.org/10.18653/v1/2021.eacl-main.121} {{BART}-{TL}:
  Weakly-supervised topic label generation}.
\newblock In \emph{Proceedings of the 16th Conference of the European Chapter
  of the Association for Computational Linguistics: Main Volume}, pages
  1418--1425, Online. Association for Computational Linguistics.

\bibitem[{Pushkarna et~al.(2022)Pushkarna, Zaldivar, and
  Kjartansson}]{10.1145/3531146.3533231}
Mahima Pushkarna, Andrew Zaldivar, and Oddur Kjartansson. 2022.
\newblock \href {https://doi.org/10.1145/3531146.3533231} {Data cards:
  Purposeful and transparent dataset documentation for responsible ai}.
\newblock In \emph{Proceedings of the 2022 ACM Conference on Fairness,
  Accountability, and Transparency}, FAccT '22, page 1776–1826, New York, NY,
  USA. Association for Computing Machinery.

\bibitem[{Rahimi et~al.(2023)Rahimi, Hoover, Mimno, Naacke, Constantin, and
  Amann}]{rahimi2023contextualized}
Hamed Rahimi, Jacob~Louis Hoover, David Mimno, Hubert Naacke, Camelia
  Constantin, and Bernd Amann. 2023.
\newblock \href {http://arxiv.org/abs/2305.14587} {Contextualized topic
  coherence metrics}.

\bibitem[{Rahimi et~al.(2024)Rahimi, Mimno, Hoover, Naacke, Constantin, and
  Amann}]{DBLP:conf/eacl/RahimiMHNCA24}
Hamed Rahimi, David Mimno, Jacob~Louis Hoover, Hubert Naacke, Cam{\'{e}}lia
  Constantin, and Bernd Amann. 2024.
\newblock \href {https://aclanthology.org/2024.findings-eacl.123}
  {Contextualized topic coherence metrics}.
\newblock In \emph{Findings of the Association for Computational Linguistics:
  {EACL} 2024, St. Julian's, Malta, March 17-22, 2024}, pages 1760--1773.
  Association for Computational Linguistics.

\bibitem[{Rhody(2012)}]{rhody2012topic}
Lisa~M Rhody. 2012.
\newblock Topic modeling and figurative language.
\newblock \emph{Journal of Digital Humanities}, 2(1).

\bibitem[{Rijcken et~al.(2023)Rijcken, Scheepers, Zervanou, Spruit, Mosteiro,
  and Kaymak}]{rijcken2023towards}
Emil Rijcken, Floortje Scheepers, Kalliopi Zervanou, Marco Spruit, Pablo
  Mosteiro, and Uzay Kaymak. 2023.
\newblock Towards interpreting topic models with chatgpt.
\newblock In \emph{The 20th World Congress of the International Fuzzy Systems
  Association}.

\bibitem[{Robert and Casella(2009)}]{Robert_and_Casella_2009}
Christian~P. Robert and George Casella. 2009.
\newblock \emph{Introducing Monte Carlo Methods with R (Use R)}, 1st edition.
\newblock Springer-Verlag, Berlin, Heidelberg.

\bibitem[{Roberts et~al.(2014)Roberts, Stewart, Tingley, Lucas, Leder-Luis,
  Gadarian, Albertson, and Rand}]{https://doi.org/10.1111/ajps.12103}
Margaret~E. Roberts, Brandon~M. Stewart, Dustin Tingley, Christopher Lucas,
  Jetson Leder-Luis, Shana~Kushner Gadarian, Bethany Albertson, and David~G.
  Rand. 2014.
\newblock \href {https://doi.org/https://doi.org/10.1111/ajps.12103}
  {{Structural Topic Models for Open-Ended Survey Responses}}.
\newblock \emph{American Journal of Political Science}, 58(4):1064--1082.

\bibitem[{Robertson(2008)}]{10.1145/1390334.1390453}
Stephen Robertson. 2008.
\newblock \href {https://doi.org/10.1145/1390334.1390453} {A new interpretation
  of average precision}.
\newblock In \emph{Proceedings of the 31st Annual International ACM SIGIR
  Conference on Research and Development in Information Retrieval}, SIGIR '08,
  page 689–690, New York, NY, USA. Association for Computing Machinery.

\bibitem[{R{\"{o}}der et~al.(2015)R{\"{o}}der, Both, and
  Hinneburg}]{DBLP:conf/wsdm/RoderBH15}
Michael R{\"{o}}der, Andreas Both, and Alexander Hinneburg. 2015.
\newblock \href {https://doi.org/10.1145/2684822.2685324} {Exploring the space
  of topic coherence measures}.
\newblock In \emph{Proceedings of the Eighth {ACM} International Conference on
  Web Search and Data Mining, {WSDM} 2015, Shanghai, China, February 2-6,
  2015}, pages 399--408. {ACM}.

\bibitem[{Rohatgi et~al.(2023)Rohatgi, Qin, Aw, Unnithan, and
  Kan}]{DBLP:conf/emnlp/RohatgiQAUK23}
Shaurya Rohatgi, Yanxia Qin, Benjamin Aw, Niranjana Unnithan, and Min{-}Yen
  Kan. 2023.
\newblock \href {https://aclanthology.org/2023.emnlp-main.640} {The {ACL} {OCL}
  corpus: Advancing open science in computational linguistics}.
\newblock In \emph{Proceedings of the 2023 Conference on Empirical Methods in
  Natural Language Processing, {EMNLP} 2023, Singapore, December 6-10, 2023},
  pages 10348--10361. Association for Computational Linguistics.

\bibitem[{Santy et~al.(2023)Santy, Liang, Le~Bras, Reinecke, and
  Sap}]{santy-etal-2023-nlpositionality}
Sebastin Santy, Jenny Liang, Ronan Le~Bras, Katharina Reinecke, and Maarten
  Sap. 2023.
\newblock \href {https://doi.org/10.18653/v1/2023.acl-long.505}
  {{NLP}ositionality: Characterizing design biases of datasets and models}.
\newblock In \emph{Proceedings of the 61st Annual Meeting of the Association
  for Computational Linguistics (Volume 1: Long Papers)}, pages 9080--9102,
  Toronto, Canada. Association for Computational Linguistics.

\bibitem[{Sen(1990)}]{Sen1990}
Amartya Sen. 1990.
\newblock \href {https://doi.org/10.1007/978-1-349-20568-4_28} {Rational
  behaviour}.
\newblock In John Eatwell, Murray Milgate, and Peter Newman, editors,
  \emph{Utility and Probability}, pages 198--216. Palgrave Macmillan UK,
  London.

\bibitem[{Simon(1955)}]{15ae9893-24f8-3286-b772-e96b33422e95}
Herbert~A. Simon. 1955.
\newblock \href {http://www.jstor.org/stable/1884852} {{A Behavioral Model of
  Rational Choice}}.
\newblock \emph{The Quarterly Journal of Economics}, 69(1):99--118.

\bibitem[{Simon(1956)}]{1957-01985-001}
Herbert~A. Simon. 1956.
\newblock \href {https://doi.org/10.1037/h0042769} {{Rational choice and the
  structure of the environment.}}
\newblock \emph{Psychological Review}, 63(2):129--138.

\bibitem[{Stammbach et~al.(2023)Stammbach, Zouhar, Hoyle, Sachan, and
  Ash}]{stammbach2023revisiting}
Dominik Stammbach, Vil{\'e}m Zouhar, Alexander Hoyle, Mrinmaya Sachan, and
  Elliott Ash. 2023.
\newblock \href {https://doi.org/10.18653/v1/2023.emnlp-main.581} {{Revisiting
  Automated Topic Model Evaluation with Large Language Models}}.
\newblock In \emph{Proceedings of the 2023 Conference on Empirical Methods in
  Natural Language Processing}, pages 9348--9357, Singapore. Association for
  Computational Linguistics.

\bibitem[{Sutherland et~al.(2015)Sutherland, Cimpian, Leslie, and
  Gelman}]{https://doi.org/10.1111/cogs.12189}
Shelbie~L. Sutherland, Andrei Cimpian, Sarah-Jane Leslie, and Susan~A. Gelman.
  2015.
\newblock \href {https://doi.org/https://doi.org/10.1111/cogs.12189} {{Memory
  Errors Reveal a Bias to Spontaneously Generalize to Categories}}.
\newblock \emph{Cognitive Science}, 39(5):1021--1046.

\bibitem[{Tversky and Kahneman(1973)}]{TVERSKY1973207}
Amos Tversky and Daniel Kahneman. 1973.
\newblock \href {https://doi.org/https://doi.org/10.1016/0010-0285(73)90033-9}
  {Availability: A heuristic for judging frequency and probability}.
\newblock \emph{Cognitive Psychology}, 5(2):207--232.

\bibitem[{Tversky and Kahneman(1974)}]{Tversky_Kahneman_Judgment}
Amos Tversky and Daniel Kahneman. 1974.
\newblock \href {https://doi.org/10.1126/science.185.4157.1124} {{Judgment
  under Uncertainty: Heuristics and Biases}}.
\newblock \emph{Science}, 185(4157):1124--1131.

\bibitem[{Tversky and Kahneman(1981)}]{framing_Tversky_Kahneman}
Amos Tversky and Daniel Kahneman. 1981.
\newblock \href {https://doi.org/10.1126/science.7455683} {{The Framing of
  Decisions and the Psychology of Choice}}.
\newblock \emph{Science}, 211(4481):453--458.

\bibitem[{Törnberg and Törnberg(2016)}]{doi:10.1177/0957926516634546}
Anton Törnberg and Petter Törnberg. 2016.
\newblock \href {https://doi.org/10.1177/0957926516634546} {{Combining CDA and
  topic modeling: Analyzing discursive connections between Islamophobia and
  anti-feminism on an online forum}}.
\newblock \emph{Discourse \& Society}, 27(4):401--422.

\bibitem[{Vafa et~al.(2020)Vafa, Naidu, and Blei}]{DBLP:conf/acl/VafaNB20}
Keyon Vafa, Suresh Naidu, and David~M. Blei. 2020.
\newblock \href {https://doi.org/10.18653/V1/2020.ACL-MAIN.475} {Text-based
  ideal points}.
\newblock In \emph{Proceedings of the 58th Annual Meeting of the Association
  for Computational Linguistics, {ACL} 2020, Online, July 5-10, 2020}, pages
  5345--5357. Association for Computational Linguistics.

\bibitem[{Vul et~al.(2014)Vul, Goodman, Griffiths, and
  Tenenbaum}]{Edward_one_and_done_2014}
Edward Vul, Noah Goodman, Thomas~L. Griffiths, and Joshua~B. Tenenbaum. 2014.
\newblock \href {https://doi.org/https://doi.org/10.1111/cogs.12101} {{One and
  Done? Optimal Decisions From Very Few Samples}}.
\newblock \emph{Cognitive Science}, 38(4):599--637.

\bibitem[{Wang et~al.(2023)Wang, Díaz, Parrish, Aroyo, Homan, Serapio-García,
  Prabhakaran, and Taylor}]{Wang_2023}
Ding Wang, Mark Díaz, Alicia Parrish, Lora Aroyo, Chris Homan, Greg
  Serapio-García, Vinodkumar Prabhakaran, and Alex Taylor. 2023.
\newblock \href
  {https://research.google/pubs/all-that-agrees-is-not-gold-evaluating-ground-truth-labels-and-dialogue-content-for-safety/}
  {All that agrees is not gold: Evaluating ground truth labels and dialogue
  content for safety}.

\bibitem[{Weber(2009)}]{Weber_2009}
Lynn Weber. 2009.
\newblock \emph{Understanding Race, Class, Gender, and Sexuality}, {Second}
  edition.
\newblock Oxford University Press.

\bibitem[{{William Revelle}(2025)}]{PsyCh_Library}
{William Revelle}. 2025.
\newblock \href {https://CRAN.R-project.org/package=psych} {\emph{psych:
  Procedures for Psychological, Psychometric, and Personality Research}}.
\newblock Northwestern University, Evanston, Illinois.
\newblock R package version 2.5.3.

\bibitem[{Ying et~al.(2022)Ying, Montgomery, and
  Stewart}]{Ying_Montgomery_Stewart_2022}
Luwei Ying, Jacob~M. Montgomery, and Brandon~M. Stewart. 2022.
\newblock \href {https://doi.org/10.1017/pan.2021.33} {{Topics, Concepts, and
  Measurement: A Crowdsourced Procedure for Validating Topics as Measures}}.
\newblock \emph{Political Analysis}, 30(4):570–589.

\bibitem[{Zade et~al.(2018)Zade, Drouhard, Chinh, Gan, and
  Aragon}]{DBLP:conf/chi/ZadeDCGA18}
Himanshu Zade, Margaret Drouhard, Bonnie Chinh, Lu~Gan, and Cecilia~R. Aragon.
  2018.
\newblock \href {https://doi.org/10.1145/3173574.3173733} {Conceptualizing
  disagreement in qualitative coding}.
\newblock In \emph{Proceedings of the 2018 {CHI} Conference on Human Factors in
  Computing Systems, {CHI} 2018, Montreal, QC, Canada, April 21-26, 2018}, page
  159. {ACM}.

\bibitem[{Zangerle and Bauer(2022)}]{10.1145/3556536}
Eva Zangerle and Christine Bauer. 2022.
\newblock \href {https://doi.org/10.1145/3556536} {{Evaluating Recommender
  Systems: Survey and Framework}}.
\newblock \emph{ACM Comput. Surv.}, 55(8).

\end{thebibliography}
\bibliographystyle{acl_natbib}
\newpage
    \appendix

\section{Pre-processing}\label{app_pre_processing}
We tokenized texts using the Spacy library with model ``en\_core\_web\_sm''\footnote{\url{https://spacy.io/models/en}}. We discarded documents with less than five words. We used~\citet{DBLP:conf/nips/HoyleGHPBR21}'s library\footnote{\url{https://github.com/ahoho/topics}} to do the pre-processing. Following~\cite{DBLP:conf/acl/VafaNB20}, we considered all the unigrams appearing in at least 0.1\% and at most 30\% documents of the corpus. After pre-processing, the SENATE dataset had 17,573, the DESGIN dataset had 174,416 and the ACL dataset had 71,736 documents.

\section{Prolific Recruitment Details}\label{app_rec_details}
\subsection{SENATE dataset}\label{app_rec_senate_details}
All the users were located in the USA, with an approval rate of 95-100, and at least 10 previous submissions with political spectrum: \textit{Conservative, Moderate, Liberal}.
\subsection{DESIGN Dataset}\label{app_rec_design_details}
Users with an approval rate of 95-100, and at least 10 previous submissions with exposure to one of the following industries:\newline
\textit{Computer and Electronics Manufacturing, Research laboratories, Software, Telecommunications, Video Games}, \newline 
and with knowledge of one or more of the following software development techniques:\newline 
\textit{Cloud computing, Shell scripting, Background processing, Search technologies, Monitoring, caching, Version control, Virtualisation, Debugging, Functional testing, Unit testing, Web servers, Database management, Responsive design, UI design, A/B testing, UX}.

\section{Task Details}\label{app_task_details}
For the SENATE dataset, we provided the following warning message: \newline
\texttt{\small You may come across content that may be offensive or distressing. You can withdraw from the study by using ``Exit'' option at the bottom of the page.}

The following descriptions of datasets were provided to the users:
\begin{enumerate}[leftmargin=*]\small
    \item \textbf{ACL}: \texttt{These topics are inferred on scholarly research articles on the study of computational linguistics and natural language processing}
    \item \textbf{SENATE}: \texttt{These topics are inferred on speeches in the 114th session of Congress (2015-2017) recorded in the bound and daily editions of the United States Congressional Record.}
    \item \textbf{DESIGN}: \texttt{These topics are inferred on Stack Overflow posts regarding ``Software Design''.}
\end{enumerate}

\section{LLM based Coherence Rating}\label{app_llm_coherence}
We use the coherence rating prompts from \citet{stammbach2023revisiting} and modify them for our datasets. We use the following Likert scales:

\subsection{Likert Scales}
Let $K$ denote the number of items in a Likert scale:
\begin{enumerate}[leftmargin=*,topsep=0pt]\itemsep0pt
    \item $K$ = 3 \newline
    \textbf{LS-3}: ``1'' = \textit{not very related}, ``2'' = \textit{moderately related}, ``3'' = \textit{very related}  
    \item $K$ = 5 \newline
    \textbf{LS-5}: ``1'' = \textit{not very related}, ``2'' = \textit{somewhat related}, ``3'' = \textit{moderately related}, ``4'' = \textit{related}, ``5'' = \textit{very related}  
\end{enumerate}

\subsubsection{System prompts \textit{with} a task and dataset description}
\begin{prompt}\label{prompt_acl_senate_coherence}\small
You are a helpful assistant evaluating the top words of a topic model output for a given topic. 
    
Please rate how related the following words are to each other on a scale from 1 to \{$\mathbf{K}$\} (\{\textbf{LS-}$\mathbf{K}$\}). 
    
[dataset\_details]

Reply with a single number, indicating the overall appropriateness of the topic.
\end{prompt}
\noindent where $K\in{3,5}$.

\subsubsection{Task and dataset details}
\begin{itemize}[leftmargin=*,topsep=0pt]\itemsep0pt \small
    \item \textbf{ACL}:  
    \texttt{The topic modeling is based on the ACL Anthology corpus. The corpus consists of scholarly research articles on the study of computational linguistics and natural language processing.}
    \item \textbf{SENATE}:  
    \texttt{The topic modeling is based on the United States Congressional Record. The corpus consists of speeches from the 114th session of Congress (2015-2017) spoken on the floor of each chamber of Congress: the United States House of Representatives and the United States Senate.}
    \item \textbf{DESIGN}:
    \texttt{The topic modeling is based on Stack Overflow posts regarding ``Software Design''.}
\end{itemize}

\subsubsection{System prompts \textit{without} a task and dataset description}
\begin{prompt}\label{prompt_coherence}\small
    You are a helpful assistant evaluating the top words of a topic model output for a given topic. 
    
    Please rate how related the following words are to each other on a scale from 1 to \{$\mathbf{K}$\} (\{\textbf{LS-}$\mathbf{K}$\}). 
    
    Reply with a single number, indicating the overall appropriateness of the topic.
    
\end{prompt}

\noindent where $K\in{3,5}$.

\subsubsection{Implementation}
We use the same hyperparameter settings\footnote{\url{https://github.com/dominiksinsaarland/evaluating-topic-model-output/blob/main/src-human-correlations/chatGPT_evaluate_topic_ratings.py}} as that of \citet{stammbach2023revisiting} while using GPT. We prompt GPT (GPT-3.5-Turbo) five times and report the average score.

\section{Topic-Construct Agreement}\label{app_sec_error_plots}
\begin{landscape}
\begin{figure}
  \centering
\includegraphics[scale=0.625]{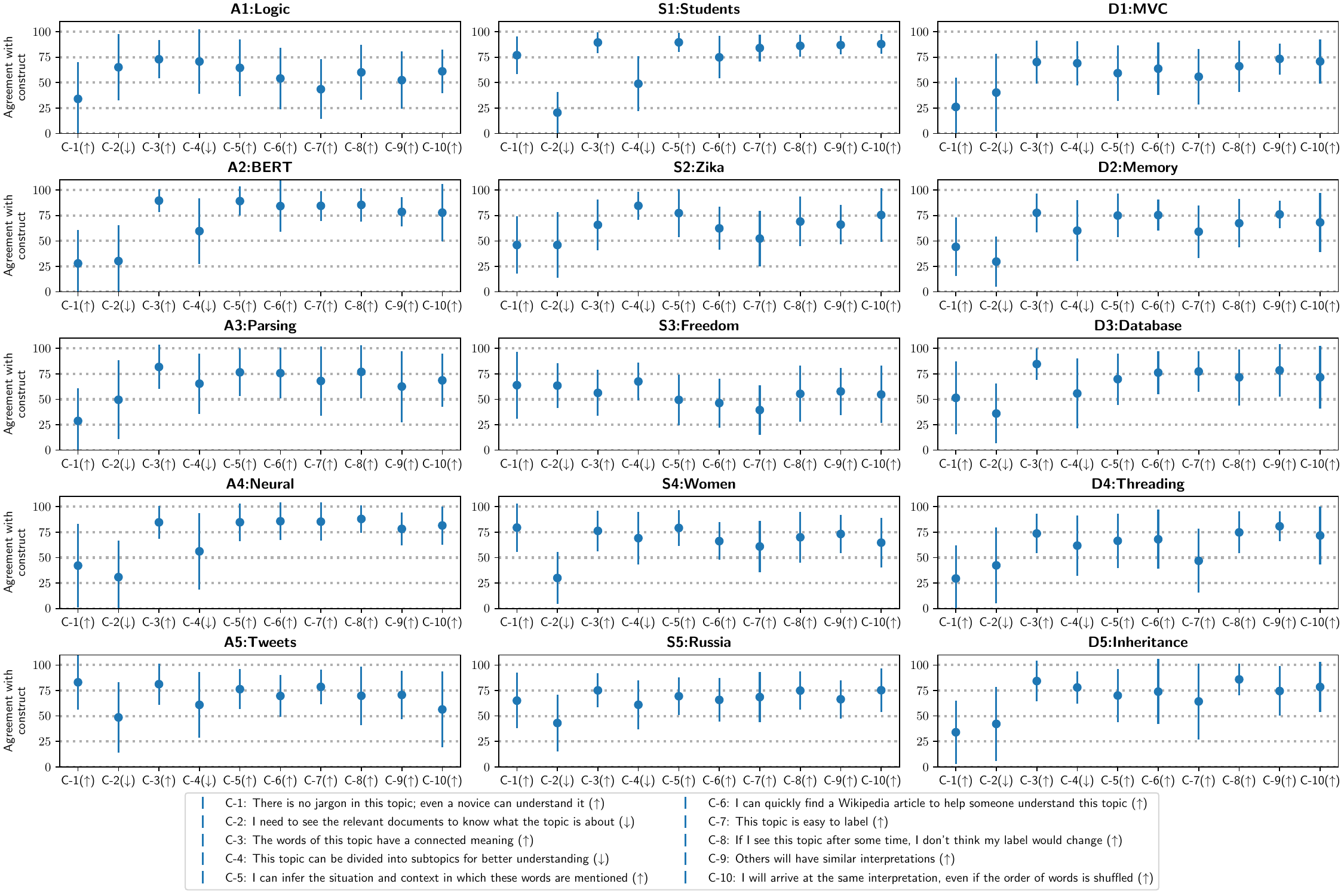} 
%    \caption{Caption place holder}
\caption{Mean and the coefficient of variation of agreement of users for each topic and construct pair.}
\label{fig_acl_senate_plots_app}
\end{figure}
\end{landscape}
\onecolumn
\section{Example Topic Card}\label{app_example_topic_card}
Table~\ref{tab_app_topic_card} provides a Topic Card by a fictional interpreter.
\begin{table*}[h!]\small
\begin{tabular}{|p{\linewidth}|}
\hline
Topic 32 ($T_W$): \texttt{shared code subtask hindi identification ensemble submitted team teams detection participants subtasks submission mixed official track character semeval submissions run preprocessing tamil runs competition validation} \tabularnewline
\begin{itemize}
    \item The task, purpose, goals, or research questions intended to address or achieve using topics \newline 
    \texttt{Develop a classification scheme for organizing research articles related to Computational Linguistics}
    \item User positionality \newline 
    \texttt{I am a third-year female Ph.D. student at the University of XXX, USA. My research interests lie at the intersection of Natural Language Processing and Social Science. I completed my undergraduate education in Electrical Engineering from Indian Institute XXX in 2020. I am proficient in Bengali, Hindi, and English. I have basic proficiency in French. I am a citizen of India.}
    \item Rationale behind topic interpretation and label for all the users \newline 
    \texttt{The topic is about ``Shared Tasks for NLP in Indian Languages''. There are many words (`shared', `subtask', `track', `semeval', etc.) about shared tasks. However, the topic has the word ``hindi'' in the 4th position and ``tamil'' in the 22nd position. Hindi and Tamil are widely spoken languages in India; hence I think related documents will be of participation of various teams in shared tasks on Indian Languages. I could not find one reference that will discuss different shared tasks, but I am pretty confident about my interpretation.}
    \item Utility of the topic \newline
    \texttt{A person interested in Indian languages can find the topic useful.}
    \item Details of prompts if LLMs are used while interpreting topics \newline 
    \texttt{I tried the following prompts with ChatGPT}:
    \begin{itemize}
        \begin{tt}
        \item \textbf{Prompt}: Please provide a topic label that best describes this group of words: [$T_W$] \newline 
            \textbf{ChatGPT}: The best topic label for this group of words is:

``Natural Language Processing Competition''

This label captures the context of language processing (e.g., Hindi, Tamil), competition (e.g., SemEval), and the various terms associated with participation and technical aspects (e.g., submission, preprocessing, validation, ensemble).
        \item \textbf{Prompt}: Please find an intruder word from this group of words: [$T_W$].  \newline
\textbf{ChatGPT}: The intruder word in this group is ``hindi''.
\end{tt}
\end{itemize}
    \texttt{ChatGPT interpretation is \textbf{not} aligned with my interpretation}
    \item How sensitive are the end results with the interpretations\newline 
    \texttt{As I am from India, I know about Hindi and Tamil. If someone finds these two words as outliers or intruders they will arrive at different interpretations and it can affect the classification of related documents.}    
\end{itemize} 
\tabularnewline \tabularnewline \hline
\end{tabular}
\caption{A Topic Card for a topic by a fictional interpreter}
\label{tab_app_topic_card}
\end{table*}
\onecolumn
\section{Topic Labels}\label{app_topic_labels}
\begin{table*}[h]\footnotesize
\begin{tabular}{lp{0.15\linewidth}p{0.15\linewidth}p{0.15\linewidth}p{0.15\linewidth}p{0.15\linewidth}}
\toprule
 \textbf{Topic} $\rightarrow$ & \textbf{\ref{top_acl_1}}         & \textbf{\ref{top_acl_2}}                      & \textbf{\ref{top_acl_3}} & \textbf{\ref{top_acl_4}}                & \textbf{\ref{top_acl_5}}        \\ %\midrule
 \textbf{User} $\downarrow$ &&&&& \\ \midrule
AI-0          & reasoning, inference, logical    & transformer, bert, fine tuning                & parser, tree,treebank    & neural network, embedding, architecture & social media, user messages     \\ \midrule
AI-1          & theory                           & pretrained language model                     & Parse tree               & deep learning                           & social platform                 \\ \midrule
AI-2          & science                          & language model                                & semantics                & deep learning                           & social media                    \\ \midrule
AI-3          & language                         & LLM                                           & data                     & attention                               & zombie                          \\ \midrule
AI-4          & Natural language inference       & BERT                                          & dependency parser        & neural network                          & Social media (Twitter)          \\ \midrule
AI-5          & natural language inference       & fine-tuning pre-trained large language models & dependency parsing       & deep neural network architecture        & social media                    \\ \midrule
AI-6          & Entailment and Logical Reasoning & Transformer-based Encoder Models              & Syntactic Parsing        & Deep Learning                           & Social Media Platform - Twitter \\ \midrule
AI-7          & proof                            & pre-trained language model                    & parse tree               & deep learning                           & social media                    \\ \midrule
AI-8          & Logical entailment               & BERT                                          & Dependency parsing       & Neural Networks                         & Twitter/X                       \\ \midrule
AI-9          & Logic                            & Natural Language Processing                   & "parser"                 & Artificial Intelligence                 & social media                    \\ \midrule
AI-10         & predictive modeling              & Large language models                         & linguistic               & deep learning models                    & social media                    \\ \midrule
AI-11         & Entailment Task                  & Transformer Language Models                   & Dependency Parsing       & Neural Network                          & Social Media Tools              \\ \bottomrule
\end{tabular}
\caption{Labels of ACL dataset topics provided by users}
\label{tab_acl_labels}
\end{table*}

\begin{table*}[h!]\footnotesize
\begin{tabular}{lp{0.15\linewidth}p{0.15\linewidth}p{0.15\linewidth}p{0.15\linewidth}p{0.15\linewidth}}
\toprule
 \textbf{Topic} $\rightarrow$ & \textbf{\ref{top_senate_1}}         & \textbf{\ref{top_senate_2}}                      & \textbf{\ref{top_senate_3}} & \textbf{\ref{top_senate_4}}                & \textbf{\ref{top_senate_5}}        \\ %\midrule
 \textbf{User} $\downarrow$ &&&&& \\ \midrule
SI-0          & Financial aid and student loan debt                          & Public funding for women's reproductive   health related to the Zika virus & Religious freedom                               & Equal rights for women                  & National security                                   \\ \midrule
SI-1          & student loan debt                                            & Healthcare of pregnant women                                               & Religious freedom in America and the world      & Equal rights of women in America        & Unites States foreign relations                     \\ \midrule
SI-2          & Student loan cancellation                                    & Abortion and birth control                                                 & The cuban missile crisis in America.            & equal rights in the workplace           & International security between NATO and its allies. \\ \midrule
SI-3          & issue of student loan debt and cost of education             & US funding for public health                                               & role of freedom of speech and religious freedom & social movements and advocacy in the US & the state of US relations with foreign governments  \\ \midrule
SI-4          & Financial Aid for College Students                           & Pregnancy Issues in the United States                                      & Differences between Cuba and the United States  & American Human Rights                   & Ukraine and Russian Conflict                        \\ \midrule
SI-5          & Student loan discussions                                     & African Aid                                                                & Talk about Cuban lifting embargo                & Women's rights                          & International relations                             \\ \midrule
SI-6          & Student Loan Forgiveness                                     & Zika women's health access                                                 & American Greatness                              & Inequality seen in America              & National Security and International Relations       \\ \midrule
SI-7          & Newly graduated college students and financial aid and debt. & The health of pregnant women and epidemics.                                & Religious freedom in the United States.         & Pay equality for women and minorities.  & The United States role in foreign policy.           \\ \midrule
SI-8          & Desirable vs. Undesirable                                    & Health Emergencies                                                         & Random                                          & equal opportunity                       & Assistance                                          \\ \midrule
SI-9          & Student loans and financial aid                              & Healthcare                                                                 & International influence                         & Equal Rights                            & Foreign policy with Russia                          \\ \midrule
SI-10         & Student loans                                                & Healthcare access                                                          & Freedom of religion in Cuba                     & Civil rights                            & Foreign policy towards Russia and Ukraine           \\ \midrule
SI-11         & Students loan and state programs                             & Zika virus and women pregrenancy or abortion                               & freedom or religion                             & vote rights and discrimination          & US and Ukraine conflict with russia and China       \\  \midrule
SI-12         & University Student Loans and Financial Aid                   & Health Issues (Infectious Diseases and Abortion)                           & Religious Freedom in American Government        & Equal Rights for Women                  & Foreign Policy                                      \\ \midrule
SI-13         & Federal student loan repayment                               & Zika virus and pregnancy                                                   & Spanish-American War                            & Feminist Movement                       & International relations  \\ \bottomrule                          
\end{tabular}
\caption{Labels of SENATE dataset topics provided by users}
\label{tab_senate_labels}
\end{table*}

\begin{table*}[h!]\footnotesize
\begin{tabular}{lp{0.15\linewidth}p{0.15\linewidth}p{0.15\linewidth}p{0.15\linewidth}p{0.15\linewidth}}
\toprule
 \textbf{Topic} $\rightarrow$ & \textbf{\ref{top_design_1}}         & \textbf{\ref{top_design_2}}                      & \textbf{\ref{top_design_3}} & \textbf{\ref{top_design_4}}                & \textbf{\ref{top_design_5}}        \\ %\midrule
 \textbf{User} $\downarrow$ &&&&& \\ \midrule
DI-0          & Software architecture and development   frameworks      & Computer memory, performance, and caching      & Database management (using SQL)                 & Multithreading                                                      & Object oriented programming and inheritance                               \\ \midrule
DI-1          & Design patterns                                         & Memory management                              & SQL database                                    & multithreading                                                      & Inheritance in OOP                                                        \\ \midrule
DI-2          & Software arhitectual patterns such as MVC, MVVM and MVP & Performance optimization and memory management & Creating, labeling and designing SQL databases. & Concurrency control, database managment and transaction processing. & Object oriented programming concepts, such as how class inheritance works \\ \midrule
DI-3          & MVC related questions                                   & Chaching data to improve performance           & Understanding database operations and queries   & Using threads correctly                                             & Defining classes and subclasses                                           \\ \midrule
DI-4          & Design Patterns for UI dev                              & Computer memory and performance                & Relational databases and SQL                    & Concurrency and Processes Management                                & Object-oriented Programming OOP                                           \\ \midrule
DI-5          & Web Development                                         & Memory and performance optimazation            & Database management with SQL                    & Multithreaded programming                                           & Object-oriented programming                                               \\ \midrule
DI-6          & Web development                                         & Computer Architecture                          & MySQL                                           & Operating Systems                                                   & Object oriented programming                                               \\ \midrule
DI-7          & MVC Architecture                                        & Database Performance Optimization              & Database Management Commands                    & Concurrency Control in Multi-Threaded Systems                       & OOP                                                                       \\ \midrule
DI-8          & software architecture MVC/MVM                           & Performance optimization                       & SQL                                             & multi-threading                                                     & object-oriented programming                                               \\ \midrule
DI-9          & Introduction to Object-Oriented Programming Concepts    & Software Performance Optimization              & Database Optimization                           & Locking and Transaction Management in Computer Programming          & Object-Oriented Programming Concepts                                      \\ \bottomrule     
\end{tabular}
\caption{Labels of DESIGN dataset topics provided by users}
\label{tab_design_labels}
\end{table*}
\onecolumn
\section{Rationales of Topic Interpretations}
\begin{table*}[h!]\footnotesize
\begin{tabular}{P{1.25cm}P{2.5cm}P{11cm}}
\toprule
\textbf{ID} & \textbf{Topic label} & \textbf{Rationale} \\ \midrule 
\multicolumn{3}{p{14cm}}{\textit{\textbf{A. Familiarity effects for AI-10}}} \\ \midrule
Topic \ref{top_acl_2}  & 	Large language models &	the topic is easy to understand what it is "about", as it seems to capture relevant keywords to LLMs. Seems the terms convey a cohesive topic, I wouldn't change my labeling even if I spend more time looking at these terms. That said, providing top documents could change labeling and make it more specific. Topic can be dividied to sub topics, such as modeling, data preporcessing for llms, etc. \\ \midrule
Topic \ref{top_acl_3} &	{linguistic} & 	this topic for sure is less familar to me. looking at top terms sicj as parser, tree, head, dependecy, I said ok it is about linguistic. but going further down the list the terms became less familiar to me. so perhaps if I would look down the list first (the order was suffled) I would arrive at different answers. I think dividing it to two topics would make one topic that is so familar to me (lingusitics) and another one that I wouldn't be familiar to me (Penn, attachment, projective, constituency) \\ \midrule\midrule
\multicolumn{3}{p{14cm}}{\textit{\textbf{B. Context effect: Topic \ref{top_design_4}}}} \\ \midrule
{DI-6}  & Operating Systems & Threads, mutex, locking, process are big hints. A subject I had in Uni was called "Operating Systems" that also covered this topic. I'm sure there is a better label such as multi-threaded programming but I  decided to go for a more general one. \\ \midrule
DI-2  &	Concurrency control, database managment and transaction processing.&	There is plenty of jargon, a novice without at least some background knowledge wouldn't understand the topic. The situation and context depends on the order of the words, its a bit of a broad topic. The meaning of the words vary a lot but are centered around the same topic. The topic is very broad so I don't need to see any relevant documents. The topic would require more than a wikipedia article to understand, but there are plenty of other sources on the internet that help. I provided that label because the words such as "lock", "thread", "time", "multiple", and "transaction" reminded me of it the most. Its hard to label because the order of the words could shift my interpretation. The words are really specific so I don't think my label would change. I think the interpretations would vary a little but would probably be similar because the topic is so specific. If the order is shuffled I might think of another similar label. Its a complex topic, it has a couple words that are based on some subtopics (such as "read" "write" "access").\\ \midrule\midrule
\multicolumn{3}{p{14cm}}{\textit{\textbf{C. Memory based fallback effect: Topic \ref{top_acl_3}}}} \\  \midrule
{AI-3} & {data} & key word: dependency, tree, treebank
Maybe data structure? I'm not sure  \\ \midrule
{AI-9} & {parser} & Not familiar with this topic, but chose parser because I've used "parser". \\ \midrule\midrule 
\multicolumn{3}{p{14cm}}{\textit{\textbf{D. Ordering effect: Topic \ref{top_acl_1}}}} \\ \midrule 
{AI-6} & {Entailment and Logical Reasoning} &  The topic is not easy to label because there seems to be a mixture of multiple topics like entailment (NLI), logical reasoning, logic formulae, and predicates. However, entailment is a little more dominant sub-topic because of word like "entailment", "nli", "hypothesis". These words would appear as a jargon to anyone who is a novice to NLP. As there is a mixture of multiple sub-topics in this topic, others are likely to have little different interpretation and even my own label may change if I see this topic again after a long time. The order of words is not playing much role in reducing the labelling confusion of this topic. \\ \midrule
{AI-8} &{Logical\newline entailment} & The first two words (logical entailment) suggested the topic for me and the other words seemed to align. I might have gone for NLI if that term appeared at the top. \\ 
\bottomrule
\end{tabular}
\caption{Examples of topic interpretations showing the effect of differential salience of words}
\label{tab_ecological_frequency_app}
\end{table*}

\begin{table*}[h!]\footnotesize
%\begin{tabular}{P{0.8cm}P{2.cm}P{11.95cm}}
\begin{tabular}{P{1.35cm}P{2.1cm}P{11.3cm}}
\toprule
\textbf{ID} & \textbf{Topic label} & \textbf{Rationale} \\ \midrule 
%\multicolumn{3}{p{14cm}}{\textit{Topic \ref{top_senate_5}}} \\ \midrule 
{SI-4 for \textbf{\ref{top_senate_5}}} & {Ukraine and\newline  Russian\newline Conflict} & 	All the words have to do with countries that have some influence over the current conflict between Russia and Ukraine, as well as points of contention throughout the war (economic and security impacts, for example). This label does require some background knowledge on current events (so not everyone would come to the same conclusion), but given the context of the label the words seem related, easily understandable without extra context, can be subdivided (because there are so many aspects of the war that can be examined), and easily looked up on places like Wikipedia. However, without this knowledge, others might not have as easy of a time labeling hence the lower score there. I think because there are certain anchor words that hint at war like "allies" that my label is unlikely to change if I saw this topic after a period of time or if the words were shuffled. \\ \midrule
{{SI-8} for \textbf{\ref{top_senate_5}}} & Assistance & II labeled this section as assistance because ever since the Russians war against Ukraine has started the United states role has been providing assistance to Ukraine. The words that make the topic clear to me are of course Ukraine, allies, and relations. In my process of deciding whether I agreed or not I with a little bit hesitant because I'm not sure Wikipedia is the right source for this topic so I was considering not suggesting that to someone that to someone but I realized the question is asking if I could not if I should.]\\ \midrule% \midrule 
{SI-5 for \textbf{\ref{top_senate_1}}}&	{Student loan \newline discussions} & Words like students, financial, loans, and banks make me think the speech was about the recent rates of student loans and how they need to be fixed. Perhaps a bill was introduced during the speech.	 \\ \midrule
{SI-9 for \textbf{\ref{top_senate_2}}} & {Healthcare} & The words could be grouped into multiple different yet more specific labels. For example: women, abortion, planned, parenthood, womens, control, babies, pregnant. All these words remind me on the increasing national regulation on women's healthcare rights. Words: zika, virus, ebola, disease; are more related to international healthcare issues.  \\ 
\bottomrule
\end{tabular}
\caption{Example topic interpretations with presentism effect}
\label{tab_biases_app}
\end{table*}

\begin{table*}[h!]\footnotesize
\begin{tabular}{P{1.35cm}P{2.1cm}P{11.3cm}}
\toprule
\textbf{ID} & \textbf{Topic label} & \textbf{Rationale} \\ \midrule 
\multicolumn{3}{p{14cm}}{\textit{\textbf{A. Stereotypes}}} \\ \midrule
{SI-5 for\newline \textbf{\ref{top_senate_2}}} & {African Aid} & Things like, abortion, zika, ebola, are all indicative of ongoing African societal and medical issues. This speech could be about aid to countries in Africa. \\ \midrule
{SI-9 for\newline \textbf{\ref{top_senate_3}}} & {International influence} & Unclear if these words are to represent the separation of church and state (freedom, religious, religion, faith) or represent USA's involvement with spreading democracy into other countries (Cuba, freedom, history, united, states, human, liberty, war, state, free, nations, democracy, political, cuban)\\ \midrule\midrule
\multicolumn{3}{p{14cm}}{\textit{\textbf{B. Levels of Abstraction: Topic \ref{top_acl_2}}}} \\ \midrule 
{AI-8} & {BERT} & The topic seems to be specifically about training and fine-tuning large language models like bert and roberta, and seems to reflect snippets common in recent NLP papers around using pretrained large language models and fine-tuning them. The order of the words do lend credence to fine-tuning in particular, otherwise I might have considered this topic to be about large language models in particular. The fact that roberta is mentioned and transformer also appears makes it see to me that it is more about LLMs in general and not just bert, but I cannot be sure. \\ \midrule% The associated terms such as roberts and masking further confirmed my analysis. \\ \midrule
{AI-10} & {Large language models} & the topic is easy to understand what it is "about", as it seems to capture relevant keywords to LLMs. Seems the terms convey a cohesive topic, I wouldn't change my labeling even if I spend more time looking at these terms. That said, providing top documents could change labeling and make it more specific. Topic can be dividied to sub topics, such as modeling, data preporcessing for llms, etc. \\ \midrule\midrule 
\multicolumn{3}{p{14cm}}{\textit{\textbf{C. Generalization for Topic \ref{top_acl_1}}}} \\ \midrule
{AI-2} & {science} & I think a person requires some experience in the sciences to understand these terms. Even though the terms are related, they seem to be overly broad. It is very difficult to label these terms under a label that has a narrower scope. I believe that this topic can be split into a topic that encompasses the mathematical terms such as proof, formula, expressions, etc, and another topic that focuses on logical reasoning such as rule, entailment, predicate, etc. \\  \midrule
AI-9 & Logic & I can roughly guess the content because I've learned these concepts in a course, such as first-order logic, but I can't recall the name, as I used "logic" as the topic. \\ \midrule
AI-4 & Natural language inference &	I am really hesitating between logics and nli as "entailment", "semantic", and "inference" are more related to NLI while the first word is "logical" \\ \bottomrule
\end{tabular}
\caption{Examples of topic interpretations showing the effect of Stereotyping and Generalization}
\label{tab_stereotypes_app}
\end{table*}

\begin{table*}[h!]\footnotesize
%\begin{tabular}{P{1.6cm}P{3.5cm}P{9.3cm}}
\begin{tabular}{P{1.6cm}P{2.6cm}P{10.5cm}}
\toprule
\textbf{ID} & \textbf{Topic label} & \textbf{Rationale} \\ \toprule
\multicolumn{3}{p{14cm}}{\textbf{\textit{A. Gestalt principle: Closure}}} \\ \midrule 
{SI-11 for \textbf{\ref{top_senate_5}}} & {US and Ukraine conflict with russia and China} & Very hard topic to understand without the context or representative documents. It could be easily considred as the recent
war between Ukraine and Russia, but if I recall correctly the corpus is formed between 2015-2017. My guess is this topic is about
the Ukrain joining Nato struggle and conflicts with Russia with China and US being involved.
\\ \midrule
{DI-4 for \textbf{\ref{top_design_4}}} & {Concurrency and Processes Management} & For me the words: "thread", "process", "shared", "mutex", and "lock" all are related on how different processes or units of execution access the same resources avaliable. "transaction", "update", "block", and "locked" $->$ are words for me that focus more on the integrity of the data, makes me think of atomic operations\\ \midrule\midrule%, makes me think of atomic operations\\ \bottomrule
\multicolumn{3}{p{14cm}}{\textbf{\textit{B. Gestalt principle: Figure and Ground for Topic~\ref{top_senate_4}}}} \\ \midrule 
SI-4 & American Human Rights & There are words having to do with different categories of people (women, men, black), words having to do with rights (equality, amendment, fair, justice, civil), and words that are about America specifically (maryland, americans). Based on this, I decided on the label "American Human Rights" because that phrase seems to capture the idea of keeping "justice" for all the rights specified above. This label was a little tricky to decide on because of the specific words about America being interspersed with civil rights language. However, the words still seem, on a whole, related to each other given the label, and could be further researched on Wikipedia. For a better label, however, I'd need to see more related documents. Shuffling/seeing this label after a long time may possibly affect my label, especially if more America-centric words were included earlier in the shuffle. This is why I think others would have a different label. And because there are so many aspects to civil and human rights, this topic can easily be subdivided for better understanding. \\ \midrule
SI-5 & Women’s rights & Things like equality, Black, women, rights, discrimination all point to a speech about Women's rights and the struggles women have historically faced in society. \\ \midrule
SI-9 & Equal Rights & After looking at the first few words "women","rights","act","equal" all of these related towards the movement for equality between women and men. As I read further down the list, there continued to words of a similar theme (discrimination, men, fair) or repeated words (equality, womens).
 \\ \bottomrule
\end{tabular}
\caption{Topic interpretations showing effects of Gestalt principles}
\label{tab_gestalts_emergence_app}
\end{table*}

\begin{table*}[h!]\footnotesize
\begin{tabular}{P{0.8cm}P{2.cm}P{11.85cm}}
\toprule
\multicolumn{3}{p{15.5cm}}{\textit{\textbf{\ref{top_senate_3}}: freedom religious american history rights government united states americans human nation world america religion cuba liberty war state free faith nations democracy society political cuban}} \\ \midrule
\textbf{ID} & \textbf{User's Label} & \textbf{Rationale} \\ \midrule 
{SI-12} &	{Religious Freedom in American Government} &	{Given the context of the corpus (U.S. Senate speeches), this topic is fairly coherent but rather general. It contains many words about American government and politics (ex: government, united, states, america, nation, world), and also includes several words related to religion (religious, religion, faith). Since "religious" and "freedom" are the top two words, I would guess that this topic is more specifically referring to religious freedom in American government, but I'm not sure -- I would need to see documents to confirm this. Also, it isn't clear to me whether "cuba" and "cuban" belong in this topic.} \\ \midrule
{SI-13} &	{Spanish-American War} &	{Words like Cuban, America, war, and freedom indicated to me that the topic is about the Spanish-American War, which primarily revolved around Cuban independence. The words related to religion/faith threw me off the most, which led me to lowering my confidence-related scores. I'd need to see the document. I think the document might be about the role of religion in Spanish-American War, which would explain the topic better.} \\ \bottomrule 
\end{tabular}
\caption{Examples of topic interpretations based on anchoring-and-adjustment heuristic.}
\label{tab_aa_examples_app}
\end{table*}

\end{document}